\documentclass[journal]{IEEEtran}
\ifCLASSINFOpdf
\else
\fi
\usepackage{soul}
\usepackage{subfig}
\usepackage{todonotes}
\usepackage{graphicx}
\usepackage{amsmath}
\usepackage{amssymb}
\usepackage{bm}
\usepackage{bbm}
\usepackage{cite}
\usepackage[super]{nth}
\usepackage{textcomp}
\DeclareMathOperator{\tr}{Tr}
\usepackage{float}
\usepackage{gensymb}
\usepackage{enumitem}
\usepackage{algorithm,algorithmic}
\usepackage{multirow}
\usepackage[utf8]{inputenc}
\usepackage{xcolor}
\def\BibTeX{{\rm B\kern-.05em{\sc i\kern-.025em b}\kern-.08em
		T\kern-.1667em\lower.7ex\hbox{E}\kern-.125emX}}

\newcommand\nextline{\nonumber \\ &\hspace{0.4cm}}
\newcommand\Tstrutt{\rule{0pt}{1ex}}         
\newtheorem{cor}{Corollary}

\newtheorem{lem}{Lemma}

\newcommand{\E}{\mathbb{E}}

\newcommand{\blkdiag}{\operatorname{blkdiag}}

\newcommand{\N}{\mathcal{N}}

\usepackage{siunitx}

\newcommand{\diag}{\operatorname{diag}}
\renewcommand{\baselinestretch}{0.9}
\usepackage{todonotes}
\usepackage{cuted}
\allowdisplaybreaks
\setstcolor{red}
\makeatletter
\newenvironment{breakablealgorithm}
{
	\begin{center}
		\refstepcounter{algorithm}
		\hrule height.8pt depth0pt \kern2pt
		\renewcommand{\caption}[2][\relax]{
			{\raggedright\textbf{\ALG@name~\thealgorithm} ##2\par}
			\ifx\relax##1\relax 
			\addcontentsline{loa}{algorithm}{\protect\numberline{\thealgorithm}##2}
			\else 
			\addcontentsline{loa}{algorithm}{\protect\numberline{\thealgorithm}##1}
			\fi
			\kern2pt\hrule\kern2pt
		}
	}{
		\kern2pt\hrule\relax
	\end{center}
}
\makeatother

\begin{document}
	\title{Random Matrix Based Extended Target Tracking with Orientation: A New Model and Inference} 
	
	\author{Barkın~Tuncer,~\IEEEmembership{Student Member,~IEEE,} Emre~\"{O}zkan,~\IEEEmembership{Member,~IEEE,}
		
		\thanks{Authors are with the Department
			of Electrical and Electronics Engineering, Middle East Technical University, Ankara,
			Turkey  e-mail: barkint@metu.edu.tr, emreo@metu.edu.tr.}
	}
	
	\maketitle
	
	\begin{abstract}
		In this study, we propose a novel extended target tracking algorithm which is capable of representing the extent of dynamic objects as an ellipsoid with a time-varying orientation angle.
		A diagonal positive semi-definite matrix is defined to model objects' extent within the random matrix framework where the diagonal elements have inverse-Gamma priors. 
		The resulting measurement equation is non-linear in the state variables, and it is not possible to find a closed-form analytical expression for the true posterior because of the absence of conjugacy. 
		We use the variational Bayes technique to perform approximate inference, where the Kullback-Leibler divergence between the true and the approximate posterior is minimized by performing fixed-point iterations.
		The update equations are easy to implement, and the algorithm can be used in real-time tracking applications. We illustrate the performance of the method in simulations and experiments with real data. The proposed method outperforms the state-of-the-art methods when compared with respect to accuracy and robustness. 
	\end{abstract}
	
	
	\IEEEpeerreviewmaketitle

	\section{Introduction}
	\IEEEPARstart{E}{xtended} 
	target tracking (ETT) problem involves processing multiple measurements that belong to a single target at each scan. In contrast to conventional tracking algorithms, which rely on point target assumption, ETT algorithms aim at estimating the target extent, which can be defined as the target-specific region that generates multiple measurements.  Previous studies in the ETT literature can be broadly categorized into four groups:
	\begin{itemize}
		\item Simple shape models
		\item Random matrix (RM) based models
		\item Random hyper-surface (RHS) based models
		\item Mixture models
	\end{itemize}
	
	\noindent A simple approach to ETT involves assuming a predefined shape for the extent/contour of the object such as a circle, a rectangle, or a line \cite{Circle, Rect, Stick}. 
	The most common approaches in the literature utilize RM models, where the target extent is represented by an ellipse \cite{Koch, Feldman, UmutO, lan2012trackinga}. 
	Alternatively, RHS models are suggested in \cite{Baum1, Baum2}. 
	More recently, Gaussian Process (GP) based models are proposed for extended target tracking \cite{EoGp1, EoGp2, kumru2018}. 
	Another fold of studies focuses on modeling the target extent with multiple ellipses \cite{RongLi, KarlETT, QiETT,kara2018}.
	
	RM models represent the elliptical extent of a target by an unknown positive semi-definite matrix (PSDM). 
	In the Bayesian framework, inverse-Wishart (IW) distribution defines a conjugate prior for PSDMs. 
	In RM based ETT models, the overall target state is composed of a Gaussian kinematic state vector and an IW distributed extent matrix. 
	Several algorithms are proposed to approximate or compute the posterior of this augmented state. 
	In \cite{Koch}, exact inference is performed by neglecting the measurement noise and exploiting the resulting conjugacy. 
	This model is restrictive in the sense that the kinematic state vector has to be composed of the target's position and higher-order spatial components such as velocity and acceleration. 
	Koch's RM model is later improved in  \cite{Feldman} to account for the measurement noise in the updates at the expense of exact inference. 
	The update equations in \cite{Feldman} are intuitive, but the approximations are difficult to quantify theoretically. 
	This problem is later addressed by \cite{UmutO}, where the variational Bayes technique is used to obtain approximate posteriors. 
	
	None of the aforementioned RM models is capable of tracking the heading angle of an extended target. 
	They instead rely on a forgetting factor to forget the sufficient statistics of the unknown extent matrix in time to account for the changes in the orientation of the target. 
	Such an approach is problematic as it aims to discard the information collected in the past and try to explain the change in the orientation as the change in the target shape.
	There are earlier studies that aim at estimating the orientation angle of elliptical objects  \cite{yang,yang2017,yang2016,granstrom2014}.
	In \cite{granstrom2014}, the orientation of the target is estimated by using the information that is obtained from the trajectory of the target. 
	The methods that are proposed in \cite{yang,yang2017,yang2016} express the unknown extent parametrically and perform inference using extended Kalman filters together with pseudo-measurements. In these approaches, an explicit nonlinear measurement equation is derived where the kinematic and shape parameters are related to measurements by multiplicative random variables.
	The inference in \cite{yang2016} involves second-order Taylor series approximation to approximate the pseudo-measurement covariance matrix.
	In \cite{yang2017}, the authors improved the algorithm in \cite{yang2016} further and eliminated the need for computing Hessian matrices. 
	Instead, they showed that the expectation and the covariance of the pseudo-measurements could be approximated from the original measurement covariance matrix. 
	In \cite{yang}, the predicted measurement covariance matrix approximation is calculated more precisely. 
	
	There are several drawbacks of the methods in \cite{yang,yang2017,yang2016}. The models used in these methods involve a multiplicative noise term, which introduces additional non-linearity in the problem, and it makes performing inference more difficult. The methods require a pseudo-measurement, which must be constructed from the original measurements to update kinematic and extent states separately. The measurements collected at one time instant must be processed sequentially. Changing the order of the measurements causes minor changes in the performance \cite{yang}. The state variables corresponding to the semi-axes lengths, which are positive by definition, are distributed with Gaussian distributions whose support covers both positive and negative real line. In some cases, it can be challenging to reflect available information into the priors defined in \cite{yang}, which may cause a collapse in the extent estimates in the subsequent measurement updates. 
	
	In this work, we propose a novel RM model that defines a Gaussian prior for the heading angle and an inverse Gamma prior for the extent parameters, which guarantee positive semi-definiteness. It is not possible to find a closed-form expression for the resulting posterior hence we utilize the variational Bayes technique to perform approximate inference. The variational Bayes technique is successfully applied to complex filtering problems in the literature to obtain approximate posteriors \cite{smidl2008variational, sarkka2009recursive, nurminen2018skew, nurminen2015robust, huang2017novel, UmutO}.

	The contributions of this manuscript can be listed as follows.
	\begin{itemize}
		\item We provide a novel solution that can track the orientation of a target and estimate its extent jointly.
		\item The proposed solution utilizes appropriate priors, which are defined over non-negative real numbers, for the unknown extent parameters. 
		\item The problem formulation does not rely on multiplicative noise terms or pseudo-measurements. 
		\item The measurement update can be performed by processing multiple measurements as a batch. The update does not depend on the order of the measurements.
		\item The uncertainty in the orientation and shape parameters can be expressed separately.
		\item The inference is performed via the well-known variational Bayes technique. 
	\end{itemize}
	
	The paper is organized as follows. In Section \ref{sec:problem}, we formulate the problem of joint shape estimation and tracking of elliptical objects with time-varying orientation. In the subsequent sections, we present the inference method. The measurement update is derived in Section \ref{sec:measupdate}. Time update is given in Section \ref{sec:timeupdate}.
	A closer look at a single measurement update and its comparison with the state-of-the-art extended Kalman filter (EKF) algorithm is given in Section \ref{sec:single_meas}.
	Lastly, the results are presented and discussed in Section \ref{sec:results}.
	
	\begin{table}[htbp]
		\caption{NOTATIONS}
		\label{table:notation}
		\begin{center}
			\hrule \Tstrutt
			\begin{itemize}[leftmargin=*]
				\item Set of real matrices of size $m \times n$ is represented with $\mathbb{R}^{m \times n}$.
				\item Set of symmetric positive definite and semi-definite matrices of size $n \times n$ is represented with $\mathbb{S}_{++}^{n}$ and $\mathbb{S}_+^{n}$, respectively.
				
				\item $\mathcal{N}(\mathbf{x}; \boldsymbol{\mu},\Sigma)$ represents the multivariate Gaussian distributions with mean vector $\boldsymbol{\mu} \in \mathbb{R}^{n_x}$ and covariance matrix $\Sigma \in \mathbb{S}_{++}^{n_x}$,
				
				\item  $\mathcal{IG} (\sigma; \alpha,\beta)$ represents the inverse Gamma distribution over the scalar $\sigma \in \mathbb{R}^{+}$ with shape and scale parameters $\alpha \in \mathbb{R}^{+}$ and $\beta \in \mathbb{R}^{+}$ respectively,
				\begin{align*} 
				\mathcal{IG} (\sigma; \alpha,\beta) &= \frac{\beta^\alpha}{\Gamma(\alpha)}\sigma^{-\alpha-1}\exp\left(-\frac{\beta}{\sigma}\right),
				\end{align*}
				\item The number of measurements at time $k$ is represented by $m_k \in \mathcal{Z}^+$.
				\item For given measurement number of $m_k$, $\textbf{Y}_k$ represents the measurement set $\{ \mathbf{y}_k^1, \dots, \mathbf{y}_k^{m_k} \}$ at time $k$. 
				\item For any number $a \in \mathcal{Z}^+$, $\textbf{Z}_k$ represents the variable set $\{ \mathbf{z}_k^1, \dots, \mathbf{z}_k^{a} \}$ at time $k$. 
				\item $\textbf{r}_k$ represent the vector $[ r_k^1, \dots, r_k^{a} ]^T$ with size $a \in \mathcal{Z}^+$.
				\item \text{KL} denotes the Kullback-Leibler divergence between two distributions $q(x)$ and $p(x)$,
				\begin{align*}
				\text{KL} \bigl ( q(x) || p(x) \bigr )  \triangleq \int q(x) \log \left( \frac{q(x)}{p(x)} \right)dx.
				\end{align*}
				\item $\text{det}(A)$ denotes the determinant of matrix $A$. 
				\item $\tr \big[A\big] = \sum_{i=1}^{n} a_{ii}$ where $a_{ii}$ is the $i^{\text{th}}$ diagonal element of $A \in \mathbb{R}^{n \times n} $.
				\item $\E_p$ denotes the expectation operator, and $p$ emphasizes the underlying probability distribution(s).
				\item $\diag(a_1,a_2, \ldots, a_n)$ returns the diagonal matrix whose diagonal elements are $a_1, a_2, \ldots, a_n$.
				\item $\blkdiag(A_1, A_2, \ldots, A_N)$ returns the block diagonal square matrix whose main-diagonal blocks are the input matrices $A_1, A_2,\ldots, A_N$.
				\item $h.o.t.$ stands for higher-order terms.
			\end{itemize}
			\hrule
		\end{center}
	\end{table}
	
	\section{Problem Definition}
	\label{sec:problem}
\begin{figure} 
		\centering
		\includegraphics[scale=0.3]{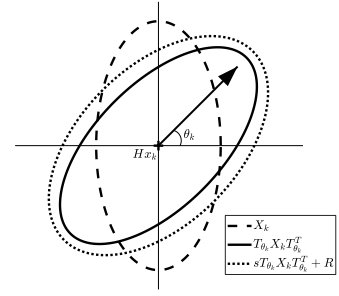}
		\caption{An illustration of the target extent model with ellipsoids corresponding to the covariance matrices $X_k$ (dashed line), $T_{\theta_k} X_k T_{\theta_k}^T$ (solid line) and  $sT_{\theta_k} X_k T_{\theta_k}^T + R$ (dotted line), respectively.}
		\label{fig:dynamic}
	\end{figure}

	Consider a single target from which multiple measurements are generated in a single scan.
	Assume that the state of the extended target consists
	of the kinematic state $\mathbf{x}_k \in \mathbb{R}^{n_x}$, the orientation angle $\theta_k \in\mathbb{R}$, and the diagonal positive definite target extent matrix $X_k\in\mathbb{R}^{n_y\times n_y}$, $X_k\triangleq \diag\left( \sigma_{k}^{1},\sigma_{k}^{2}, \ldots, \sigma_{k}^{n_y}  \right)$, where $n_x$ and $n_y$ represent the dimensions of the kinematic target state and the measurements,
	respectively. Given $\mathbf{x}_k, X_k$ and $\theta_k$, the measurements generated by the target are assumed to be independent and identically distributed,
	\begin{align}
	\label{eq:meas}
	p(\mathbf{y}_k^j\big|\mathbf{x}_k,X_k,\theta_k)\sim\N\big(\mathbf{y}_k^j;H \mathbf{x}_k,sT_{\theta_k}X_kT_{\theta_k}^T+R\big),
	\end{align}
	where
	\begin{itemize}
		\item $\mathbf{y}_k^j \in \mathbb{R}^{n_y}$ is the $\text{j}^{\text{th}}$ measurement at time $k$, 
		\item $H\in\mathbb{R}^{n_y\times n_x}$ is the measurement matrix,
		\item $R\in\mathbb{R}^{n_y\times n_y}$ is the positive definite measurement noise covariance matrix,
		\item $s\in\mathbb{R}^+$ is the scaling parameter,
		\item $T_{\theta_k}\in\mathbb{R}^{n_y\times n_y}$ is the rotation matrix which performs a rotation around the center of the target by the orientation angle $\theta_k$. $T_{\theta_k}$ satisfies the well known properties of the rotation matrices such as  $T_{\theta_k}^{-1}=T_{\theta_k}^T$, and $\text{det}(T_{\theta_k})=1$.
		In 2D, it is defined as,
		\begin{align}
		T_{\theta_k} \triangleq \begin{bmatrix}
		\cos(\theta_k) & -\sin(\theta_k)\\ 
		\sin(\theta_k) & \cos(\theta_k)
		\end{bmatrix}.
		\end{align}
	\end{itemize}
	Note that the measurement likelihood in (\ref{eq:meas}) can be interpreted as a measurement model with two additive Gaussian terms; one with time-varying, unknown but state-dependent statistics, 
	$v^I_k(X_k,\theta_k) \sim \mathcal{N}(0, sT_{\theta_k} X_k T^T_{\theta_k})$, and one with known statistics, $v^{II}_k \sim \mathcal{N}(0, R)$.  
	\begin{equation} \label{eq:yk}
	y_k = Hx_k + v^I_k(X_k,\theta_k) +v^{II}_k
	\end{equation}
	The effective covariance matrix in likelihood (\ref{eq:yk}) is unknown, time-varying, and state-dependent, which casts the main difficulty in the ETT problem together with the absence of conjugacy\footnote{A family of prior distributions is conjugate to a particular likelihood function if the posterior distribution belongs to the same family as the prior.}. An illustration of the resulting extent model is depicted in Figure \ref{fig:dynamic}.
	Similar elliptical models are frequently used in target tracking applications for tracking vehicles, vessels, pedestrians, animals, or groups of objects \cite{granstrom2017pedestrian, guerlin2020study, scheidegger2018mono, veiback2016tracking, vivone2019converted}.
	
	In the Bayesian filtering framework, we aim at estimating the unknown variables $\mathbf{x}_k, \theta_k$, and $X_k$ given the measurements collected up to and including time $k$.  To achieve this, we define appropriate priors for the unknowns and try to compute their posteriors in a recursive manner. This is generally performed by repeating two recursive steps:
	\begin{itemize}
		\item \textbf{Time Update (Prediction):} At any time step $k$, the predictive distribution  $p(\mathbf{x}_k,X_k,\theta_k|\textbf{Y}_{1:k-1})$ is computed according to Chapman-Kolmogorov equation by using the posterior from the previous time step $k-1$, and the transition density induced by the system dynamics.
		\item \textbf{Measurement Update (Correction):} When the new measurements  $\textbf{Y}_k$ are available, the posterior distribution $p(\mathbf{x}_k,X_k,\theta_k|\textbf{Y}_{1:k})$ is computed by using the Bayes' rule. In this step, the predictive distribution $p(\mathbf{x}_k,X_k,\theta_k|\textbf{Y}_{1:k-1})$ is used as the prior.
	\end{itemize}
	Unfortunately, it is not possible to obtain a closed form expression for the posterior in our problem. Therefore, we will look for an approximate analytical solution using a variational approximation.
	
	Before introducing the details of this approximation, we will first define the prior distributions of the unknown variables. The joint prior distribution of the kinematic state, the extent, and the orientation is specified as
	\begin{align}
	\label{eq:prior}
	p(\mathbf{x}_0,X_0,\theta_0) &= \mathcal{N}\left(\mathbf{x}_0; \hat{\mathbf{x}}_{0},P_{0}\right)  \times \prod_{i=1}^{n_y} \mathcal{IG}\left(\sigma_{0}^{i}; \alpha_{0}^{i}, \beta_{0}^{i}\right) \nextline \times\mathcal{N}\left(\theta_0; \hat{\theta}_{0},\Theta_{0}\right),
	\end{align}
	where $X_0\triangleq \diag\left( \sigma_{0}^{1}, \sigma_{0}^{2}, \ldots, \sigma_{0}^{n_y}  \right),$ and $\mathcal{IG}\left(\sigma_{0}^{i}; \alpha_{0}^{i}, \beta_{0}^{i}\right)$ denotes the inverse Gamma distribution.
	Here, $\hat{\mathbf{x}}_{0}$ and $P_{0}$ are the prior mean and covariance matrix of the Gaussian kinematic state vector $\hat{\mathbf{x}}_{0}$, respectively. 
	The prior mean and covariance matrix of the orientation angle ${\theta}_{0}$ are denoted by $\hat{\theta}_{0}$ and $\Theta_{0}$, respectively.
	Please see Table~\ref{table:notation} for the complete list of the notations.
	
	\noindent In the following sections, we will describe the \emph{Measurement Update} and \emph{Time Update} steps of the proposed method in detail.   
	
	\section{Measurement Update}
	\label{sec:measupdate}
	Suppose at time $k$, we have the following conditional predicted density for the kinematic, orientation, and extent states:
	\begin{align}\label{model2}
	p(\mathbf{x}_{k}, X_{k} ,\theta_k| \textbf{Y}_{1:k-1}) &= \mathcal{N}(\mathbf{x}_{k}; \hat{\mathbf{x}}_{k|k-1}, P_{k|k-1})\nextline \times \prod_{i=1}^{n_y} \mathcal{IG}(\sigma_{k|k-1}^{i}; \alpha_{k|k-1}^{i}, \beta_{k|k-1}^{i})\nextline\times \mathcal{N}\left(\theta_k; \hat{\theta}_{k|k-1},\Theta_{k|k-1}\right).
	\end{align} 
	The left-hand side of the above expression is conditioned on the measurements up to and including time instant $k-1$.
	The predicted mean and covariance of the Gaussian state vector is represented by $\hat{\mathbf{x}}_{k|k-1}$ and $P_{k|k-1}$, respectively. The shape and the scale variables for the $i^\text{th}$ diagonal element of the inverse Gamma distributed extent state $X_k$ are $\alpha^i_{k|k-1}$ and $\beta^i_{k|k-1}$.
	
	\noindent When the measurements $\textbf{Y}_{k}$ are available at time $k$, the posterior distribution can be computed using Bayes' rule
	\begin{align} \label{bayes}
	p&(\mathbf{x}_{k}, X_{k} ,\theta_k| \textbf{Y}_{1:k}) \nonumber \\
	&= \frac{p(\textbf{Y}_{k}| \mathbf{x}_{k}, X_{k} ,\theta_k ) p(\mathbf{x}_{k}, X_{k} ,\theta_k| \textbf{Y}_{1:k-1}) } { p( \textbf{Y}_{k} | \textbf{Y}_{1:k-1}) }. 
	\end{align}

	\noindent By assuming conditional independence of the measurements at time $k$, the measurement likelihood can be factorized as
	\begin{align} \label{likelihood}
	p(\textbf{Y}_{k}| \mathbf{x}_{k}, X_{k} ,\theta_k ) =& \prod_{j=1}^{m_k} p(\mathbf{y}_{k}^j | \mathbf{x}_{k}, X_{k} ,\theta_k) \nonumber \\
	=& \prod_{j=1}^{m_k} \mathcal{N}(\mathbf{y}_{k}^j; H \mathbf{x}_k, sT_{\theta_k}X_kT_{\theta_k}^T + R ).
	\end{align}
	
	In the following, we will describe a variational inference based approximation method to estimate the posterior distribution using the likelihood function in \eqref{likelihood}.
	
	\subsection{Variational Inference}
	\label{sec:variter}
	An approximate analytical solution for the posterior density in \eqref{bayes} can be obtained as a product of factorized probability density functions (PDFs) using a variational approximation. 
	Before we present the details, we need to define additional instrumental variables to address   
	the absence of conjugacy caused by
	the additive measurement noise covariance term $R$ in the likelihood. 
	We will call these variables noise-free measurements \cite{UmutO}, and denote them with  $\textbf{Z}_k = \{ \mathbf{z}_k^j \}_{j=1}^{m_k}$. 
	By using $\textbf{Z}_k$, the measurement likelihood in \eqref{likelihood} can be expressed as
	\begin{align} \label{meas_1}
	\mathcal{N}( \mathbf{y}_k^j;  H& {\mathbf{x}}_k, sT_{\theta_k}X_k T_{\theta_k}^T + R ) =\nonumber \\
	\int \mathcal{N}(\mathbf{y}_k^j; & \mathbf{z}_k^j, R)\mathcal{N}(\mathbf{z}_k^j;  H {\mathbf{x}}_k , sT_{\theta_k}X_k T_{\theta_k}^T )d\mathbf{z}_k^j
	\end{align}
	for a single measurement.
	Note that the measurement likelihood is the marginal of the following joint density
	\begin{align}
	p(\mathbf{y}_k^j, \mathbf{z}_k^j | {\mathbf{x}}_k, X_k)& =\mathcal{N}(\mathbf{y}_k^j; \mathbf{z}_k^j, R)\nextline \times \mathcal{N}(\mathbf{z}_k^j; {H}{\mathbf{x}}_k, sT_{\theta_k}X_k T_{\theta_k}^T).
	\end{align}
	Let us include the instrumental variable $\textbf{Z}_k$ in the posterior. Later, it will be marginalized out to obtain the posterior of the states
	\begin{align} \label{modified_posterior}
	p({\mathbf{x}}_k, X_k, \theta_k, \textbf{Z}_k | \textbf{Y}_{1:k} ) \approx q_\mathbf{x}(\mathbf{x}_k) q_{X} (X_k) q_{\theta}(\theta_k) q_{\textbf{Z}} (\textbf{Z}_k).
	\end{align} 
	Here, $q_{\textbf{Z}}(\textbf{Z}_k)$ denotes the approximate density of the instrumental variable $\textbf{Z}_k$. The idea of variational approximation is to seek factorized densities whose product minimizes the following cost function.
	\begin{align} \label{VarInf}
	\hat{q}_\mathbf{x}, \hat{q}_{X},\hat{q}_\theta, \hat{q}_{\textbf{Z}} =& \arg\min\limits_{q_\mathbf{x}, q_{X}, q_\theta, q_{\textbf{Z}}}  \text{KL} \bigl ( q_\mathbf{x}(\mathbf{x}_k) q_{X} (X_k) q_{\theta}(\theta_k) q_{\textbf{Z}} (\textbf{Z}_k) \nonumber \\
	& || p({\mathbf{x}}_k, X_k,\theta_k, \textbf{Z}_k | \textbf{Y}_{1:k}) \bigr ).
	\end{align}
	The solution of the optimization problem (\ref{VarInf}) satisfies the following equation \cite[Ch.~10]{Bishop}:
	\begin{align} \label{logexp}
	\log \hat{q}_\phi (\phi_k)=& \E_{\backslash\phi} \bigl[ \log p(\mathbf{x}_k, X_k,\theta_k, \textbf{Z}_k,\textbf{Y}_{k}|\textbf{Y}_{1:k-1}) \bigr] + c_{\backslash\phi}
	\end{align}
	where $\phi \in \{\mathbf{x}_k, X_k, \theta_k, \mathbf{Z}_{k} \}$, and $\backslash\phi$ is the set of all elements except $\phi$, e.g., $\E_{\backslash\mathbf{x}_k}$ will denote expectation with respect to variables $X_k$, $\theta_k$, $\mathbf{Z}_{k}$.
	The constant term with respect to variable $\phi$ will be denoted by $c_{\backslash\phi}$.
	The joint density $p(\mathbf{x}_k, X_k,\theta_k, \textbf{Z}_k, \textbf{Y}_{k}  | \textbf{Y}_{1:k-1})$ in (\ref{logexp}) can be written explicitly as
	\begin{align} \label{jointdist}
	p&(\mathbf{x}_k, X_k,\theta_k, \textbf{Z}_k,\textbf{Y}_{k}  | , \textbf{Y}_{1:k-1}) \nonumber \\  
	=&p(\textbf{Y}_{k} |\textbf{Z}_k)p(\textbf{Z}_k|\mathbf{x}_k,X_k,\theta_k) p(\mathbf{x}_k,X_k,\theta_k|\textbf{Y}_{1:k-1}) \nonumber \\
	=&\left(\prod_{j=1}^{m_k} \mathcal{N}(\mathbf{y}_k^j; \mathbf{z}_k^j,R) \right) \left(\prod_{j=1}^{m_k} \mathcal{N}(\mathbf{z}_k^j; H\mathbf{x}_k, sT_{\theta_k}X_k T_{\theta_k}^T) \right) \nonumber \\
	&\times \mathcal{N}(\mathbf{x}_k; \hat{\mathbf{x}}_{k|k-1},P_{k|k-1}) \prod_{i=1}^{n_y} \mathcal{IG}(\sigma_{k|k-1}^{i}; \alpha_{k|k-1}^{i}, \beta_{k|k-1}^{i})\nonumber \\
	&\times\mathcal{N}(\theta_k; \hat{\theta}_{k|k-1},\Theta_{k|k-1}).
	\end{align} 
	The optimization problem (\ref{VarInf}) can be solved by fixed-point iterations  \cite[Ch.~10]{Bishop}. Each iteration is performed by updating only one factorized density in (\ref{modified_posterior}) while keeping all other densities fixed to their last estimated values.
	The update equations of the approximate densities in the $(\ell+1)^{\text{th}}$ iteration will be given in the following subsections. 
	To simplify the notations,  $p\big(\mathbf{x}_k,X_k,\theta_k,\textbf{Z}_k,\textbf{Y}_k|\textbf{Y}_{1:k-1})$ is denoted as $P^k_{\mathbf{x},X,\theta,\textbf{Z},\textbf{Y}}$ in the sequel.

	\subsubsection{Computation of $q_\mathbf{x}^{(\ell+1)}(\cdot)$}
	Substituting the previous estimates of the factorized densities into equation \eqref{logexp} yields
	\begin{align}
	\log q_\mathbf{x}^{(\ell+1)}\big(\mathbf{x}_k\big)=\E_{\backslash\mathbf{x}_k}\left[\log P^k_{\mathbf{x},X,\theta,\textbf{Z},\textbf{Y}}\right]+c_{\backslash\mathbf{x}_k}.
	\end{align}
	The expectation above can be simplified as
	\begin{subequations}
		\begin{align}
		&\E_{\backslash\mathbf{x}_k}\left[\log P^k_{\mathbf{x},X,\theta,\textbf{Z},\textbf{Y}}\right] \nonumber \\  &= \E_{\backslash\mathbf{x}_k}\left[\log P\big(\textbf{Z}_k|\mathbf{x}_k,X_k,\theta_k\big)\right] \nonumber \\ &\hspace{0.4cm}+ \log \mathcal{N}(\mathbf{x}_k; \hat{\mathbf{x}}_{k|k-1},P_{k|k-1}) +c_{\backslash\mathbf{x}_k} \\
		&=\sum_{j=1}^{m_k} -0.5\tr\bigg[\big(\overline{\mathbf{z}^j_k}-H\mathbf{x}_k\big)\big(\overline{\mathbf{z}^j_k}-H\mathbf{x}_k\big)^T \nonumber \\ &\hspace{0.4cm} \times \E_{q_{X}^{(\ell)},q_{\theta}^{(\ell)}} \left[( s T_{\theta_k} X_k T_{\theta_k}^T)^{-1} \right]\bigg] \nextline +  \log \mathcal{N}(\mathbf{x}_k; \hat{\mathbf{x}}_{k|k-1},P_{k|k-1}) + c_{\backslash\mathbf{x}_k}  \\
		&= -0.5\tr\bigg[m_k\big(\overline{\mathbf{z}_k}-H\mathbf{x}_k\big)\big(\overline{\mathbf{z}_k}-H\mathbf{x}_k\big)^T \nonumber \\ &\hspace{0.4cm} \times  \E_{q_{X}^{(\ell)},q_{\theta}^{(\ell)}} \left[( s T_{\theta_k} X_k T_{\theta_k}^T)^{-1} \right]\bigg] \nextline +  \log \mathcal{N}(\mathbf{x}_k; \hat{\mathbf{x}}_{k|k-1},P_{k|k-1}) + c_{\backslash\mathbf{x}_k}  \\ 
		&= \log \mathcal{N}(\overline{\mathbf{z}_k}; H\mathbf{x}_k, \frac{\E_{q_{X}^{(\ell)},q_{\theta}^{(\ell)}} \left[( s T_{\theta_k} X_k T_{\theta_k}^T)^{-1} \right]^{-1}}{m_k}) \nextline + \log \mathcal{N}(\mathbf{x}_k; \hat{\mathbf{x}}_{k|k-1},P_{k|k-1}) + c_{\backslash\mathbf{x}_k}, \label{qxupdate_a}
		\end{align}
	\end{subequations}
	\noindent where $\overline{\mathbf{z}^j_k} \triangleq \E_{q_{\textbf{Z}}^{(\ell)}}[\mathbf{z}_k^j]$, and $\overline{\mathbf{z}}_k \triangleq  \frac{1}{m_k}  \sum_{j=1}^{m_k}  \overline{\mathbf{z}^j_k}.$ 
	It can be seen from \eqref{qxupdate_a} that $q_\mathbf{x}^{(\ell+1)}(\mathbf{x}_k)$ is a Gaussian PDF with mean vector $\hat{\mathbf{x}}_{k|k}^{(\ell+1)}$ and covariance ${P}_{k|k}^{(\ell+1)}$  
	\begin{equation}
	q_\mathbf{x}^{(\ell+1)}(\mathbf{x}_k) = \mathcal{N}( \mathbf{x}_k; \hat{\mathbf{x}}_{k|k}^{(\ell+1)}, P_{k|k}^{(\ell+1)} ),
	\end{equation}
	where
	\small
	\begin{subequations} \label{x_est}
		\begin{align}
		\hat{\mathbf{x}}_{k|k}^{(\ell+1)} &= P_{k|k}^{(\ell+1)} (P^{-1}_{k|k-1} \hat{\mathbf{x}}_{k|k-1} \nextline + m_k H^T \E_{q_{X}^{(\ell)},q_{\theta}^{(\ell)}} \left[( s T_{\theta_k} X_k T_{\theta_k}^T)^{-1} \right] \bar{\mathbf{z}}_k ), \label{eq:x_m} \\
		P_{k|k}^{(\ell+1)} &= (P^{-1}_{k|k-1} + m_k H^T \E_{q_{X}^{(\ell)},q_{\theta}^{(\ell)}} \left[( s T_{\theta_k} X_k T_{\theta_k}^T)^{-1} \right] H )^{-1}. \label{eq:x_c}
		\end{align}
	\end{subequations} 
	\normalsize
	
	\subsubsection{Computation of $q_{X}^{(\ell+1)}(\cdot)$}
	Substituting the factorized densities from the previous variational iteration into equation \eqref{logexp} yields 
	\begin{align}\label{eq:X_ell_plusone}
	\log q_X^{(\ell+1)}\big(X_k\big)=\E_{\backslash X_k}\left[\log P^k_{\mathbf{x},X,\theta,\textbf{Z},\textbf{Y}}\right]+c_{\backslash X_k}
	\end{align}
	Substituting \eqref{jointdist} into \eqref{eq:X_ell_plusone} and grouping the constant terms with respect to $X_k$ results in
	\begin{subequations}
		\label{eq:X}
		\begin{align}
		&\E_{\backslash X_k}\left[\log P^k_{\mathbf{x},X,\theta,\textbf{Z},\textbf{Y}}\right] \nonumber \\  &= \E_{\backslash X_k}\left[\log P\big(\textbf{Z}_k|\mathbf{x}_k,X_k,\theta_k\big)\right] \nonumber \\ &\hspace{0.4cm}+ \sum_{i=1}^{n_y} \log \mathcal{IG}(\sigma_{k|k-1}^{i}; \alpha_{k|k-1}^{i}, \beta_{k|k-1}^{i}) +c_{\backslash X_k} \\
		&= \frac{-m_k}{2}\log|sX_k| \nextline - \frac{1}{2} \tr \bigg[\sum_{j=1}^{m_k}\E_{\backslash X_k}\big[(\mathbf{z}^j_k-H\mathbf{x}_k)(\mathbf{z}^j_k-H\mathbf{x}_k)^T \nextline \times (sT_{\theta_k}X_kT_{\theta_k}^T)^{-1}\big]\bigg] \nextline + \sum_{i=1}^{n_y} \log \mathcal{IG}(\sigma_{k|k-1}^{i}; \alpha_{k|k-1}^{i}, \beta_{k|k-1}^{i}) + c_{\backslash X_k}.
		\end{align}
	\end{subequations}
	Consequently, the approximate posterior density $q_{X}$ follows an inverse-Gamma distribution
	\begin{equation}
	q_{X}^{(\ell+1)}(X_k) = \prod_{i=1}^{n_y} \mathcal{IG}(\sigma_{k|k}^{i,(\ell+1)}; \alpha_{k|k}^{i,(\ell+1)}, \beta_{k|k}^{i,(\ell+1)}),
	\end{equation}
	where
	\begin{subequations}\label{X_est}
		\begin{align}
		\alpha_{k|k}^{i,(\ell+1)}&=\alpha_{k|k-1}^{i}+{0.5}m_k,\label{eq:a_m}\\
		{\beta_{k|k}^{i,(\ell+1)}} &= \beta_{k|k-1}^{i}+\frac{1}{2s}\sum_{j=1}^{m_k} \E_{q_{\mathbf{x}}^{(\ell)},q_{\theta}^{(\ell)},q_{\mathbf{Z}}^{(\ell)}}[\widetilde{\mathbf{z}}^j_k (\widetilde{\mathbf{z}}^j_k)^T]_{ii}, \label{eq:b_m} 
		\end{align}
	\end{subequations}
	and $\widetilde{\mathbf{z}}_k^j \triangleq T_{\theta_k}^T(\mathbf{z}_k^j - H\mathbf{x}_k)$.
	\subsubsection{Computation of $q_{\textbf{Z}}^{(\ell+1)}(\cdot)$} 
	Substituting the factorized densities from the previous variational iteration into equation \eqref{logexp} yields 
	\begin{align}
	\log q_\textbf{Z}^{(\ell+1)}\big(\textbf{Z}_k\big)=\E_{\backslash\mathbf{Z}_k}\left[\log P^k_{\mathbf{x},X,\theta,\textbf{Z},\textbf{Y}}\right]+c_{\backslash\textbf{Z}_k}.
	\end{align}
	The expectation above can be expressed as
	\begin{subequations}
		\begin{align}
		&\E_{\backslash\mathbf{Z}_k}\left[\log P^k_{\mathbf{x},X,\theta,\textbf{Z},\textbf{Y}}\right] \nonumber \\  &= \E_{\backslash\mathbf{Z}_k}\left[\log P\big(\textbf{Z}_k|\mathbf{x}_k,X_k,\theta_k\big)\right] \nonumber \\ &\hspace{0.4cm}+ \sum_{j=1}^{m_k} \log \mathcal{N}(\mathbf{y}^j_k;\mathbf{z}^j_k,R) + c_{\backslash\textbf{Z}_k} \\
		&= \sum_{j=1}^{m_k} \frac{-1}{2} \tr\bigg[(\mathbf{z}^j_k - H\overline{\mathbf{x}_k})(\mathbf{z}^j_k - H\overline{\mathbf{x}_k})^T \nextline \times \E_{q_{X}^{(\ell)},q_{\theta}^{(\ell)}} \left[( s T_{\theta_k} X_k T_{\theta_k}^T)^{-1} \right] \bigg] \nonumber \\ &\hspace{0.4cm}+ \sum_{j=1}^{m_k} \log \mathcal{N}(\mathbf{y}^j_k;\mathbf{z}^j_k,R) + c_{\backslash\textbf{Z}_k},
		\end{align}
	\end{subequations}
	
	\noindent where $\overline{\mathbf{x}_k} = \E_{q_\mathbf{x}^{(\ell)}}[\mathbf{x}_k]$.
	Update equations for the approximate posterior density $q_{\textbf{Z}}$ in the $(\ell+1)^{\text{th}}$ iteration are given by
	\begin{equation}
	q_{\textbf{Z}}^{(\ell+1)}(\textbf{Z}_k) = \prod_{j=1}^{m_k} \mathcal{N}( \mathbf{z}_k^j; \hat{\mathbf{z}}_{k}^{j,(\ell+1)},  \Sigma_{k}^{z,(\ell+1)} ),
	\end{equation}
	\noindent where
	\begin{subequations} \label{Z_est}
		\begin{align}
		\hat{\mathbf{z}}_k^{j,(\ell+1)} &= \Sigma_k^{z,(\ell+1)} \bigg( \E_{q_{X}^{(\ell)},q_{\theta}^{(\ell)}} \left[( s T_{\theta_k} X_k T_{\theta_k}^T)^{-1} \right] H \E_{q_\mathbf{x}^{(\ell)}}[\mathbf{x}_k] \nonumber \\ &\hspace{0.4cm}+ R^{-1}\mathbf{y}_k^j \bigg), \label{eq:z_m}\\
		\Sigma_k^{z,(\ell+1)} &= \bigg( \E_{q_{X}^{(\ell)},q_{\theta}^{(\ell)}} \left[( s T_{\theta_k} X_k T_{\theta_k}^T)^{-1} \right] + R^{-1} \bigg)^{-1}.\label{eq:z_c}
		\end{align}
	\end{subequations}
	
	\subsubsection{Computation of $q_\theta^{(\ell+1)}(\cdot)$}
	\label{eq:tet_update}
	The update equations for $q_\theta^{(\ell+1)}(\cdot)$ is obtained by substituting the factorized densities from the previous variational iteration into equation \eqref{logexp} 
	\begin{align}\label{eq:theta_ell_plusone}
	\log q_\theta^{(\ell+1)}\big(\theta_k\big)=\E_{\backslash\theta_k}\left[\log P^k_{\mathbf{x},X,\theta,\textbf{Z},\textbf{Y}}\right]+c_{\backslash\theta_k}.
	\end{align}
	Substituting \eqref{jointdist} into \eqref{eq:theta_ell_plusone} and grouping the constant terms with respect to $\theta_k$ results in
	\begin{subequations}
		\label{eq:theta}
		\begin{align}
		&\E_{\backslash\theta_k}\left[\log P^k_{\mathbf{x},X,\theta,\textbf{Z},\textbf{Y}}\right] \nonumber \\ &= \E_{\backslash\theta_k}\left[\log P\big(\textbf{Z}_k|\mathbf{x}_k,X_k,\theta_k\big)\right] \nextline + \log \mathcal{N}(\theta_k;\hat{\theta}_{k|k-1},\Theta_{k|k-1}) + c_{\backslash\theta_k} \\
		&= \frac{-1}{2} \sum_{j=1}^{m_k} \E_{\backslash\theta_k} \bigg[\tr \bigr[(\mathbf{z}^j_k-H\mathbf{x}_k)(\mathbf{z}^j_k-H\mathbf{x}_k)^T\nextline \times (sT_{\theta_k}X_kT_{\theta_k}^T)^{-1}\bigr]\bigg] \nextline + \log \mathcal{N}(\theta_k;\hat{\theta}_{k|k-1},\Theta_{k|k-1}) + c_{\backslash\theta_k} \label{theta_nonlinearity}
		\end{align}
	\end{subequations}
	Unfortunately, it is not possible to obtain an exact compact form PDF for $q_\theta^{(\ell+1)}\big(\theta_k\big)$ because of the non-linearities  involved in \eqref{theta_nonlinearity}. To address this issue, we will make a first order approximation of the non-linear function $f(\theta_k) \triangleq T_{\theta_k}^T(\mathbf{z}^j_k-H\mathbf{x}_k)$ using its Taylor series expansion around $\hat{\theta}_{k|k}^{(\ell)}$,
	\begin{align}
	&f(\theta_k) = f(\hat{\theta}_{k|k}^{(\ell)}) + \nabla f(\hat{\theta}_{k|k}^{(\ell)})(\theta_k-\hat{\theta}_{k|k}^{(\ell)}) + h.o.t.,\\
	&\text{where} \quad  \nabla f(\hat{\theta}_{k|k}^{(\ell)}) \triangleq \frac{\partial f}{\partial \theta_k} \biggr \rvert_{\theta_k=\hat{\theta}_{k|k}^{(\ell)}}. \nonumber
	\end{align} 
	By plugging in the first order approximation of $f(\theta_k)$ into \eqref{theta_nonlinearity}, the expectation term can be written as 
	\begin{align}
	&\E_{\backslash\theta_k}\left[(a-b\theta_k)^T(sX)^{-1}(a-b\theta_k)\right], \nonumber   \\
	\text{where}&\nonumber \\
	&a \triangleq \big[f(\hat{\theta}_{k|k}^{(\ell)})-\nabla f(\hat{\theta}_{k|k}^{(\ell)})\hat{\theta}_{k|k}^{(\ell)}\big], \\
	&b \triangleq -\nabla f(\hat{\theta}_{k|k}^{(\ell)}).
	\end{align} 
	
	Through algebraic manipulations, $q_\theta^{(\ell+1)}(\theta_k)$ can be expressed as a Gaussian PDF with mean vector $\hat{{\theta}}_{k|k}^{(\ell+1)}$ and covariance $\Theta_{k|k}^{(\ell+1)}$,  
	\begin{equation}
	q_\theta^{(\ell+1)}(\theta_k) = \mathcal{N}( \theta_k; \hat{{\theta}}_{k|k}^{(\ell+1)}, \Theta_{k|k}^{(\ell+1)}),
	\end{equation}
	\noindent where 
	\begin{subequations} \label{tet_est}
		\begin{align}
		\hat{\theta}^{(\ell+1)}_{k|k} &= \Theta_{k|k}^{(\ell+1)}\big( \Theta_{k|k-1}^{-1}\hat{\theta}_{k|k-1}+\delta \big),\label{eq:tet_m}\\
		\Theta_{k|k}^{(\ell+1)} &= \big (\Theta_{k|k-1}^{-1} + \Delta \big)^{-1}, \label{eq:tet_c}
		\end{align}
		\begin{align}
		\delta &=\sum_{j=1}^{m_k} \tr\bigg[\overline{sX_k^{-1}}(T_{\hat{\theta}^{(\ell)}_{k|k}}^{\prime})^T\overline{\big(\mathbf{z}_k^j-H\mathbf{x}_k\big)\big(\cdot\big)^T}(T_{\hat{\theta}^{(\ell)}_{k|k}}^{\prime})\hat{\theta}^{(\ell)}_{k|k}   \bigg] \nextline -\tr\bigg[\overline{sX_k^{-1}}T_{\hat{\theta}^{(\ell)}_{k|k}}^T\overline{\big(\mathbf{z}_k^j-H\mathbf{x}_k\big)\big(\cdot\big)^T}(T_{\hat{\theta}^{(\ell)}_{k|k}}^{\prime})   \bigg],\\
		\Delta &= \sum_{j=1}^{m_k} \tr\bigg[\overline{sX_k^{-1}}(T_{\hat{\theta}^{(\ell)}_{k|k}}^{\prime})^T\overline{\big(\mathbf{z}_k^j-H\mathbf{x}_k\big)\big(\cdot\big)^T}(T_{\hat{\theta}^{(\ell)}_{k|k}}^{\prime})\bigg], 
		\end{align}
	\end{subequations}
	\noindent where $\overline{sX_k^{-1}} = \E_{q_{X}^{(\ell)}}[(sX_k)^{-1}]$, $\overline{\big(\mathbf{z}_k^j-H\mathbf{x}_k\big)\big(\cdot\big)^T} = \E_{q_\textbf{Z}^{(\ell)},q_\textbf{x}^{(\ell)}}[\big(\mathbf{z}_k^j-H\mathbf{x}_k\big)\big(\cdot\big)^T]$, and $T_{\hat{\theta}^{(\ell)}_{k|k}}^{\prime} \triangleq \frac{\partial T_{\theta_{k}}}{\partial \theta_k} \biggr \rvert_{\theta_k=\hat{\theta}^{(\ell)}_{k|k}}$.
	
	\noindent The derivations of $\delta$ and $\Delta$ are given in Appendix \ref{appen:mM}.
	
	By using the expressions derived so far, we can set up variational iterations to find the approximate posteriors $q_\mathbf{x}$, $q_{X}$, $q_\theta$, and $q_{\textbf{Z}}$.
	The noise-free measurement set $\textbf{Z}_k$ can be marginalized out from the joint density, and an approximation for $p(\mathbf{{x}}_k, X_k, \theta_k| , \textbf{Y}_{1:k} )$ is obtained.
	
	\subsubsection{Expectation Calculations}
	\label{sec:expectation}
	The relevant expectations in the variational iterations can be computed by using the following set of equations: 
	\begin{subequations}\label{eq:expp}
		\begin{align} 
		\E&_{q_\mathbf{x}^{(\ell)}}[\mathbf{x}_k] = \hat{\mathbf{x}}^{(\ell)}_{k|k}, \\
		\E&_{q_{\textbf{Z}}^{(\ell)}}[\mathbf{z}_k^j] = \hat{\mathbf{z}}^{j,(\ell)}_{k},  \\ 
		\E&_{q_{X}^{(\ell)}}[(sX_k)^{-1}] = \diag\left( \frac{\alpha^{1,\ell}}{s\beta^{
				1,\ell}}, \frac{\alpha^{2,\ell}}{s\beta^{2,\ell}}, \ldots, \frac{\alpha^{n_y,\ell}}{s\beta^{n_y,\ell}}  \right),  \\
		\E&_{q_\textbf{Z}^{(\ell)},q_\textbf{x}^{(\ell)}}[\big(\mathbf{z}_k^j-H\mathbf{x}_k\big)\big(\cdot\big)^T] = H  P_{k|k}^{(\ell)}H^T + \Sigma_k^{z,({\ell})} \nonumber \\
		&+\left( \hat{\mathbf{z}}^{j,(\ell)}_k - H \hat{\mathbf{x}}_{k|k}^{(\ell)} \right) \left(\hat{\mathbf{z}}^{j,(\ell)}_k - H \hat{\mathbf{x}}_{k|k}^{(\ell)} \right)^T,\\
		\E&_{q_{\mathbf{x}}^{(\ell)},q_{\theta}^{(\ell)},q_{\mathbf{Z}}^{(\ell)}}[\widetilde{\mathbf{z}}^j_k (\widetilde{\mathbf{z}}^j_k)^T] \nonumber\\
		&=\E_{q_{\theta}^{(\ell)}} \biggl[ T_{\theta_k}^T\bigg(  \left( \hat{\mathbf{z}}^{j,(\ell)}_k - H \hat{\mathbf{x}}_{k|k}^{(\ell)} \right) \left(\hat{\mathbf{z}}^{j,(\ell)}_k - H \hat{\mathbf{x}}_{k|k}^{(\ell)} \right)^T  \nonumber\\
		&+H  P_{k|k}^{(\ell)}H^T + \Sigma_k^{z,(\ell)}\bigg) T_{\theta_k} \biggr], \label{tzzt} 
		\end{align}
	\end{subequations}
	where the expectation in \eqref{tzzt} can be calculated by using the identity $ T_{\theta_k}^T =  T_{-\theta_k}$ and  \textit{Lemma} \ref{lem:exact}.
	\begin{align}
	\E&_{q_{X}^{(\ell)}}[(sX_k)] = \nonumber \\ & \diag\left( \frac{s\beta^{
			1,\ell}}{(\alpha^{1,\ell}-1)}, \frac{s\beta^{2,\ell}}{(\alpha^{2,\ell}-1)}, \ldots,  \frac{s\beta^{n_y,\ell}}{(\alpha^{n_y,\ell}-1)}  \right)  \label{eq:expX}
	\end{align}
	The initial conditions for the quantities can be chosen as $\hat{\mathbf{z}}_k^{j,(0)}=\mathbf{y}_k^j$, $\Sigma_k^{z,(0)}=\E_{q_{X}^{(0)}}[(sX_k)]$, $\hat{\mathbf{x}}_{k|k}^{(0)}=\hat{\mathbf{x}}_{k|k-1}$, $P_{k|k}^{(0)}=P_{k|k-1}$, $\alpha_{k|k}^{(0)}=\alpha_{k|k-1}$ and $\beta_{k|k}^{(0)}=\beta_{k|k-1}$.
	
	\subsubsection{Calculation of $\E_{q_{X}^{(\ell)},q_{\theta}^{(\ell)}} \left[( T_{\theta_k} X_k T_{\theta_k}^T)^{-1} \right]$}
	\label{sec:expectation2}
	This expectation can be calculated exactly, thanks to the factorized distributions.
	\begin{lem}\label{lem:exact}
		Given 
		\begin{align*}
		M^{-1} = \begin{bmatrix}
		m_{11} & m_{12}\\ 
		m_{21} & m_{22}
		\end{bmatrix},
		\end{align*}
		and $q^{(\ell)}_\theta(\theta_k) = \mathcal{N}(\theta_k,\hat{\theta}^{(\ell)}_{k|k},\Theta^{(\ell)}_{k|k})$, the entries of the matrix   $\E_{q_{\theta}^{(\ell)}} \left[( T_{\theta_k} M T_{\theta_k}^T)^{-1} \right]$ can be computed as: 
		\begin{subequations}\label{exact_solution}
			\begin{align}
			\E_{q_{\theta}^{(\ell)}}& \big[( T_{\theta_k} M T_{\theta_k}^T)^{-1} \big]_{11} \nonumber \\ &= \left[ \begin{matrix}m_{11}&m_{22}&-(m_{12} +m_{21})\end{matrix} \right] K\left(\hat{\theta}_{k|k}^{(\ell)},\Theta_{k|k}^{(\ell)}\right), \\
			\E_{q_{\theta}^{(\ell)}}& \big[( T_{\theta_k} M T_{\theta_k}^T)^{-1} \big]_{12} \nonumber \\ &= \left[ \begin{matrix}m_{12}&-m_{21}&m_{11} -m_{22}\end{matrix} \right] K\left(\hat{\theta}_{k|k}^{(\ell)},\Theta_{k|k}^{(\ell)}\right), \\
			\E_{q_{\theta}^{(\ell)}}& \big[( T_{\theta_k} M T_{\theta_k}^T)^{-1} \big]_{21} \nonumber \\ &= \left[ \begin{matrix}m_{21}&-m_{12}&m_{11} -m_{22}\end{matrix} \right] K\left(\hat{\theta}_{k|k}^{(\ell)},\Theta_{k|k}^{(\ell)}\right), \\
			\E_{q_{\theta}^{(\ell)}}& \big[( T_{\theta_k} M T_{\theta_k}^T)^{-1} \big]_{22} \nonumber \\ &= \left[ \begin{matrix}m_{22}&m_{11}&m_{12} +m_{21}\end{matrix} \right] K\left(\hat{\theta}_{k|k}^{(\ell)},\Theta_{k|k}^{(\ell)}\right), 
			\end{align}
			\noindent where
			\begin{align}
			K\left(\hat{\theta}_{k|k}^{(\ell)},\Theta_{k|k}^{(\ell)}\right) &\triangleq \left[ \begin{matrix}1+\cos(2\hat{\theta}_{k|k}^{(\ell)})\exp(-2\Theta_{k|k}^{(\ell)})\\1-\cos(2\hat{\theta}_{k|k}^{(\ell)})\exp(-2\Theta_{k|k}^{(\ell)})\\\sin(2\hat{\theta}_{k|k}^{(\ell)})\exp(-2\Theta_{k|k}^{(\ell)})\end{matrix} \right].
			\end{align}
		\end{subequations}
	\end{lem} 
	\noindent
	The proof is given in Appendix \ref{appen:proof}. 
	\begin{cor}
		\begin{align} \label{eq:cor}
		&\E_{q_{X}^{(\ell)},q_{\theta}^{(\ell)}} \left[( s T_{\theta_k} X_k T_{\theta_k}^T)^{-1} \right]  \nonumber\\
		&= (1-\exp(-2\Theta_{k|k}^{(\ell)})) \frac{\tr(\E_{q_{X}^{(\ell)}}[(sX_k)^{-1}])}{2} \mathbb{I}_2 \nextline + \exp(-2\Theta_{k|k}^{(\ell)})  \left(T_{\hat{\theta}_{k|k}^{(\ell)}} \E_{q_{X}^{(\ell)}}[(sX_k)^{-1}] T_{\hat{\theta}_{k|k}^{(\ell)}}^T\right),
		\end{align}
		where $\mathbb{I}_2$ is 2x2 identity matrix. This expression is obtained from \textit{Lemma} \ref{lem:exact} by exploiting the fact that the matrix $X_k$ is diagonal by definition.
	\end{cor}
	
	A summary of the resulting iterative measurement update procedure is given in Algorithm~\ref{alg:measurementupdate}.
	
	\begin{breakablealgorithm}\label{alg:measurementupdate}
		\algsetup{linenosize=\small}
		\small
		\caption{Variational measurement update}
		\begin{algorithmic}[0]
			\STATE Given $\hat{\mathbf{x}}_{k|k-1}$, $P_{k|k-1}$, $\{\alpha_{k|k-1}^i, \beta_{k|k-1}^i\}_{i=1}^{n_y}$, $\hat{\theta}_{k|k-1}$, $\Theta_{k|k-1}$ and $\textbf{Y}_k$; calculate $\hat{\mathbf{x}}_{k|k}$, $P_{k|k}$, $\{\alpha_{k|k}^i, \beta_{k|k}^i\}_{i=1}^{n_y}$, $\hat{\theta}_{k|k}$, $\Theta_{k|k}$ as follows.
			\STATE \underline{\textbf{Initialization}}
			\STATE $\hat{\mathbf{x}}_{k|k}^{(0)}\gets\hat{\mathbf{x}}_{k|k-1}$, \quad $P_{k|k}^{(0)}\gets P_{k|k-1}$,
			\STATE $\hat{\theta}_{k|k}^{(0)} \gets \hat{\theta}_{k|k-1}$,
			\quad $\Theta_{k|k}^{(0)} \gets \hat{\Theta}_{k|k-1}$,
			\STATE $\alpha_{k|k}^{i,(0)}\gets \alpha_{k|k-1}^{i}$,
			\quad $\beta_{k|k}^{i,(0)}\gets \beta_{k|k-1}^{i}$ for $i = 1, \ldots , n_y$,
			\STATE $\mathbf{z}_{k|k}^{j,(0)}\gets \mathbf{y}_k^j$  for  {$j=1,\ldots, m_k$},
			\STATE $\Sigma_k^{z,(0)}\gets \E_{q_{X}^{(0)}}[(sX_k)] \; \; \text{using} \; \; \eqref{eq:expX}$
			\STATE \underline{\textbf{Iterations}:}
			\FOR {$\ell = 0,\ldots, \ell_{max}-1$}
			\small
			\STATE Calculate the expectations in \eqref{eq:expp}, and \eqref{eq:cor}
			\STATE Update $\hat{\mathbf{x}}_{k|k}^{(\ell+1)}$, and $P_{k|k}^{(\ell+1)}$ using \eqref{x_est}
			\STATE Update $\hat{\theta}^{(\ell+1)}_{k|k}$, and $\Theta_{k|k}^{(\ell+1)}$ using \eqref{tet_est}
			\STATE Update  $\alpha_{k|k}^{i,(\ell+1)}$, and ${\beta_{k|k}^{i,(\ell+1)}}$ using \eqref{X_est} for $i = 1, \ldots, n_y$
			\STATE Update $\hat{\mathbf{z}}_k^{j,(\ell+1)}$, and $\Sigma_k^{z,(\ell+1)}$ using \eqref{Z_est} for  {$j=1,\ldots, m_k$} 
			\ENDFOR
			\STATE \underline{\textbf{Set final estimates}:}
			\STATE $\hat{\mathbf{x}}_{k|k}=\hat{\mathbf{x}}_{k|k}^{(\ell_{\max})}$,
			\quad $P_{k|k}=P_{k|k}^{(\ell_{\max})}$,
			\STATE $\hat{\theta}_{k|k}=\hat{\theta}_{k|k}^{(\ell_{\max})}$,
			\quad $\Theta_{k|k}=\Theta_{k|k}^{(\ell_{\max})}$,
			\small
			\STATE $\alpha_{k|k}^{i}=\alpha_{k|k}^{i,(\ell_{\max})}$,
			\quad $\beta_{k|k}^{i}=\beta_{k|k}^{i,(\ell_{\max})}$ for $i = 1, \ldots, n_y$
		\end{algorithmic}
	\end{breakablealgorithm}
	
	\section{Time Update}
	\label{sec:timeupdate}
	Once the measurement update is performed, the sufficient statistics of the posterior density must be propagated in time in accordance with the target dynamics. 
	An optimal time update step requires the solution to the following Chapman-Kolmogorov equation 
	\begin{align}
	\label{eq:chap1}
	p(\mathbf{x}^a_{k}, X_{k} | \textbf{Y}_{1:k-1}) &= \int p(\mathbf{x}^a_{k}, X_{k} |\mathbf{x}^a_{k-1}, X_{k-1})\nextline p(\mathbf{x}^a_{k-1} ,X_{k-1}| \textbf{Y}_{1:k-1})d\mathbf{x}^a_{k-1}dX_{k-1},
	\end{align}
	\noindent where $\mathbf{x}^a_k \triangleq \begin{bmatrix}\mathbf{x}_k^T \, \theta_k  \end{bmatrix}^T$.
	Unfortunately, it is not possible to obtain an exact compact form analytical expression for most extended target tracking models. Therefore various independence conditions are implied to perform time updates in the literature \cite{Koch, Feldman, lan2012trackinga,granstrom2014}. For a detailed analysis of possible time update approaches, interested readers can refer to \cite{granstrom2014} and the references therein.

	In the random matrix framework, it is possible to assume that the dynamical models of the kinematic state and the extent state are independent \cite{Feldman},
	\begin{align}\label{eq:time_indep}
	p(\mathbf{x}_{k}, X_{k} | \mathbf{x}_{k-1},X_{k-1}) &= p(\mathbf{x}_{k} | \mathbf{x}_{k-1})p(X_{k} | X_{k-1}).
	\end{align}
	Consequently, the time update of the kinematic state and the extent state can be decoupled for factorised posteriors. The time update of the kinematic state follows the Kalman filter prediction equations if the underlying dynamics are linear.
	Consider the following state space model which describes the dynamics of the augmented state vector $\mathbf{x}^a_k$, 
	\begin{align} \label{dyn}
	\mathbf{x}^a_{k} &= F \mathbf{x}^a_{k-1} + u_k, \quad u_k \sim \mathcal{N}(0,Q).
	\end{align}
	The prediction density $\mathcal{N}(\mathbf{x}^a_{k|k-1}; \hat{\mathbf{x}}^a_{k|k-1}, P^a_{k|k-1})$ is obtained by updating the sufficient statistics (mean and covariance) of the Gaussian components in accordance with the system dynamics
	\begin{subequations} \label{eq:timeup}
		\begin{align}
		\hat{\mathbf{x}}^a_{k|k-1} &= {F} \hat{\mathbf{x}}^a_{k-1|k-1},\\
		{P}^a_{k|k-1} &= {F} {P}^a_{k-1|k-1} {F}^T + {Q}.
		\end{align}
	\end{subequations}
	where $ P^a_k \triangleq \blkdiag(P_k, \; \;\Theta_k)$.
	
	In most tracking applications, the exact dynamics of the extent state is unknown. Even in the case where the dynamic equations of the extent states are available, the transition density induced by the known dynamics may not lead to a prediction update that results in the same family of probability distributions using \eqref{eq:chap1}.
	If the dynamics of the extent state is slowly varying but unknown, it is possible to obtain the maximum entropy prediction density of the extent states by utilizing a forgetting factor  \cite[Theorem~1]{Mapf}.  
	In that case, the sufficient statistics of the  inverse Gamma distribution is updated as
	\begin{subequations} \label{eq:timeup2}
		\begin{align}
		\alpha^{i}_{k|k-1} &= \gamma_k \alpha^{i}_{k-1|k-1},\\
		\beta^{i}_{k|k-1} &= \gamma_k \beta^{i}_{k-1|k-1}, \quad \text{for} \quad i=1, \ldots, n_y
		\end{align}
	\end{subequations}
	where $\gamma$ is the forgetting factor. We prefer to use the maximum entropy prediction density in the time update. However, it is possible to perform alternative time updates within the proposed framework.
	
	\begin{figure}[h]
		\centering
		\includegraphics[scale=0.6,trim={0 0 0 0},clip]{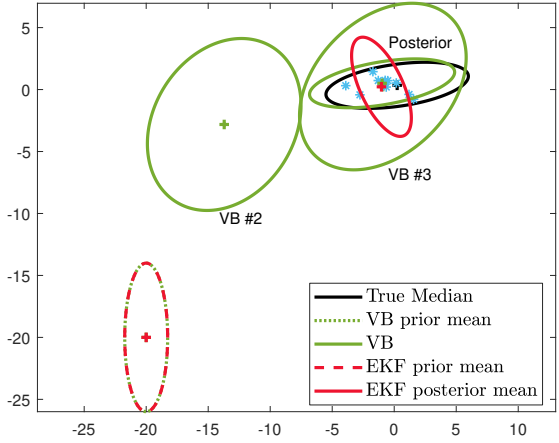}
		\captionof{figure}{A single measurement update for VB and the EKF approach. The prior and posterior mean shape estimates are represented by green dotted and solid lines for VB, respectively. The red dashed line indicates the prior mean shape estimate while the red solid line depicts the posterior mean estimate for EKF approach. The VB \#$i$ denotes the $i^{th}$ variational iteration shape estimate mean of the VB algorithm.}
		\label{fig:results2}
	\end{figure}

	\section{A Closer Look to a Single Measurement Update}
	\label{sec:single_meas}
	In this section, we investigate the proposed measurement update, here and after denoted as VB, in more detail and illustrate its capabilities in comparison with a state-of-the-art extended Kalman filter (EKF) algorithm \cite{yang}. 
	For this purpose, we initiate the prior mean and covariance of both approaches the same; and we compare the posterior distribution of the extent states.
	Consider the example given in Figure~\ref{fig:results2}, where the prior mean of the target's location is $[-20 -20]^T$. 
	The measurements are shown with blue stars, and the posterior means of the VB and EKF updates are shown with the solid green and red lines, respectively.
	The median of the true posterior, which is computed by using 1 million Monte Carlo samples, is shown with the solid black line. 
	The mean of the extent and kinematic state distributions at the end of each VB iteration is denoted by VB \#$i$ where \#$i$ stands for the $i^{th}$ variational iteration.
	A total of $10$ iterations are performed within the variational update.
	As shown in Figure~\ref{fig:results2}, the posterior found by the VB algorithm is closer to the true posterior than the posterior computed by the EKF, thanks to the iterative nature of the VB updates.
	Unlike EKF, the VB algorithm performs multiple iterations in a single update and performs multiple linearizations during the iterations by taking all available measurements into account. The ability to compute the posterior iteratively is the key concept to explain the superior performance of VB in the experiments given in Section \ref{sec:results}. 
	
	Lastly, we compare the average computation time of the algorithms. The simulations for the illustrative example are run in Matlab(R) R2019b on a standard laptop with an Intel(R) Core(TM) i7-6700HQ 2.60 GHz platform with 16 GB of RAM. We compare naive implementations of the algorithms without exploiting any code optimization methods. A single measurement update (with 10 iterations) and a single variational iteration for VB takes $2.3\times10^{-3}$ \si{sec} and $2.1\times10^{-4}$ \si{sec}, respectively.  On the other hand, it takes $8.6\times10^{-4}$ \si{sec} to perform a measurement update for the EKF. 
	The relevant parameters of the illustrative example are given in the Appendix-\ref{sec:params}.
	
	\section{Experimental Results}
	\label{sec:results}
	In this section, we evaluate the performance of the proposed method and compare it with relevant elliptical object tracking algorithms in the literature. 
	The comparison is performed through both simulations and real data experiments. 
	The alternative models are selected as the state-of-the-art EKF approach that is capable of tracking the orientation of elliptical objects  \cite{yang} and the widely used RM based ETT model \cite{Feldman}. In the sequel, we denote these algorithms as Algorithm-1 and Algorithm-2, respectively. 
	The simulation results are presented in Section~\ref{subsec: simulations}, and the results of the real-data experiment are given in Section~\ref{subsec: realdata}.

	\subsection{Simulations} \label{subsec: simulations}
	In the simulations, we use the Gaussian Wasserstein (GW) distance \cite{givens1984class}, \cite{yang2016metrics} and root-mean-square-error (RMSE) for performance evaluation and comparison,
	\begin{align}
	\label{eq:gw}
	\text{GW}&(\mathbf{m}_a,X_a,\mathbf{m}_b,X_b)^2\nonumber \\& \triangleq \underbrace{\left\lVert \mathbf{m}_a - \mathbf{m}_b \right\rVert^2_2}_{\nth{1}\;Term} +  \underbrace{\tr[X_a + X_b - 2(X_a^{\frac{1}{2}}X_bX_a^{\frac{1}{2}})^\frac{1}{2}].}_{\nth{2}\;Term}
	\end{align}
	Here, $\mathbf{m}_a$, $\mathbf{m}_b$ and $X_a$, $X_b$ stand for two different center locations and elliptic extent matrices, respectively. The first term in \eqref{eq:gw} corresponds to the error in the estimation of the object's center, and the second term corresponds to the error in extent estimation. We report both terms in \eqref{eq:gw} in addition to the overall GW distance to provide insight into the estimation performance of the algorithms in detail.
	Furthermore, we compare the RMSE of the orientation estimations which is defined by
	\begin{align}
	\label{eq:rmse}
	\text{RMSE}(\theta_{true}, \theta) = \sqrt{\frac{1}{N}\sum_{k=1}^N (\theta_{k,true} - \theta_k)^2},
	\end{align}
	\noindent where $N$ denotes the number of time steps in a single run.
	
	\subsubsection{Constant Velocity Model}
	\label{sec:cv_model}
	In the first experiment, a dynamic object is simulated, which moves according to the nearly constant velocity model defined by the following parameters.
	\begin{subequations}
		\begin{align}
		\overline{F} &= \begin{bmatrix}
		1 & T\\ 
		0  & 1
		\end{bmatrix}\otimes \mathbb{I}_2, \quad
		F = \blkdiag(\overline{F},1), \\
		P_0 &= \mathbb{I}_{5}, \quad
		\bm{\hat{\mathbf{x}}_0}=\begin{bmatrix}
		0 &  0&  50&  0& 0
		\end{bmatrix}^T, \\
		X_{true} &= \begin{bmatrix}
		50 & 0\\ 
		0 & 600
		\end{bmatrix}, \,
		\overline{Q} = \sigma^2\begin{bmatrix}
		\frac{T^3}{3} & \frac{T^2}{2}\\ 
		\frac{T^2}{2}  & T
		\end{bmatrix}\otimes \mathbb{I}_{2},\\ 
		Q &= \blkdiag(\overline{Q}, \sigma_\theta),\quad
		R = 5 \times \mathbb{I}_{2},
		\end{align}
	\end{subequations}
	\noindent where $T = 0.1$, $\sigma = 1$, and $\sigma_\theta = 0.01$.
	In this simulation, the parameters of the motion model are fully provided to the tracking algorithms so that the error due to model-mismatch does not affect the estimation performance. 
	Throughout the trajectory, the object generates an average of 10 measurements per scan. 
	We investigate two different cases separately; in the first case, the measurements follow a Gaussian distribution, and in the second case, they follow a uniform distribution. 
	All simulation experiments were performed 100 times with different realizations of the process noise, measurement noise, and measurement origin at each simulation. 
	The presented numbers are the average of these 100 Monte Carlo (MC) runs.
	The algorithm specific initial shape variables for VB are set to $\mathbf{\alpha}_0^{1,2} = [2 \; 2]^T$ and $\mathbf{\beta}_0^{1,2} = [100 \; 100]^T$. The number of variational iterations is 10.
	
	The shape variables are initialized for Algorithm-2 as $v_0 = 4$ and $V_0= \diag([100 , \; 100])$. 
	The forgetting factor is set to $\gamma = 0.99$ for both VB and Algorithm-2.
	To be consistent with \cite{yang}, we use the same notations for the parameters of Algorithm-1. 
	The prior mean and covariance matrix of the shape variables of Algorithm-1 are selected to be $\hat{\textbf{p}}_0 = [0 \; \; 10 \; \; 10]^T$ and $C^p_0 = \diag([1 , \; 20 , \; 20])$. 
	The vector $\hat{\textbf{p}}_0$ consists of $ [\theta, \; l_1, \; l_2]$ where, $\theta$, $l_1$, and $l_2$ are the orientation and the semi-axis lengths, respectively. 
	The process noise covariance matrix for the shape variables for Algorithm-1 is $C^w_p = \diag([10^{-2}, \; 0.1 , \; 0.1 ])$. 
	The kinematic state transition matrix for Algorithm-1 is set to $A_r = F(1:4,\;1:4)$.
	The initial mean of the kinematic state vector is the same as VB, $\hat{\textbf{r}}_0 = \hat{\textbf{x}}_0(1:4)$.
	The state transition matrix for the shape variables is $A_p = \mathbb{I}_{3}$.
	The initial values of the shape and kinematic variables are selected to make the prior means of the algorithms the same.
	The algorithmic specific parameters are hand-tuned to obtain the best performance of each algorithm.
	We report the average GW distance and the orientation RMSE for Gaussian and uniformly distributed measurements in Table~\ref{table:gaussian_results_gw} and Table~\ref{table:uniform_results_gw}, respectively.
	The proposed algorithm performs better in terms of estimating the extent of the target and outperforms the other algorithms in terms of GW distance.
	Additionally, VB shows a better performance in estimating the orientation of the target. 
	\begin{table}
		\centering
		\caption{The GW distance values for Gaussian measurements.}
		\label{table:gaussian_results_gw}
		\begin{tabular}{|c|c|c|c|c|}
			\hline
			\multirow{2}{*}{ } & \multirow{2}{*}{\textbf{\begin{tabular}[c]{@{}c@{}}GW Distance \\ \nth{1} Term {[}m\textsuperscript{2}{]}\end{tabular}}} &\multirow{2}{*}{\textbf{\begin{tabular}[c]{@{}c@{}}GW Distance \\ \nth{2} Term {[}m\textsuperscript{2}{]}\end{tabular}} } &  \multicolumn{2}{c|}{\textbf{\begin{tabular}[c]{@{}c@{}}GW \\ Distance {[}m{]}\end{tabular}}} \\ \cline{4-5}
			&  & & \textbf{Mean} & \multicolumn{1}{c|}{\textbf{Std.}} \\ \hline
			\textbf{Algorithm-1}& 4.78& 8.54& 3.18 &0.24 \\ \hline
			\textbf{Algorithm-2} & 5.22& 56.63& 6.86 & 1.68\\ \hline
			\textbf{VB}& \textbf{4.49}& \textbf{5.27}& \textbf{2.85}& 0.23\\ \hline
		\end{tabular}
	\end{table}
	
	\begin{table}
		\centering
		\caption{The GW distance values for uniformly distributed measurements.}
		\label{table:uniform_results_gw}
		\begin{tabular}{|c|c|c|c|c|}
			\hline
			\multirow{2}{*}{ } & \multirow{2}{*}{\textbf{\begin{tabular}[c]{@{}c@{}}GW Distance \\ \nth{1} Term {[}m\textsuperscript{2}{]}\end{tabular}}} &\multirow{2}{*}{\textbf{\begin{tabular}[c]{@{}c@{}}GW Distance \\ \nth{2} Term {[}m\textsuperscript{2}{]}\end{tabular}} } &  \multicolumn{2}{c|}{\textbf{\begin{tabular}[c]{@{}c@{}}GW \\ Distance {[}m{]}\end{tabular}}} \\ \cline{4-5}
			&  & & \textbf{Mean} & \multicolumn{1}{c|}{\textbf{Std.}} \\ \hline
			\textbf{Algorithm-1}& 1.93& 6.02& 2.49 & 0.55\\ \hline
			\textbf{Algorithm-2} & 2.18& 58.54& 6.69 & 1.65\\ \hline
			\textbf{VB}& \textbf{1.92}& \textbf{4.54}& \textbf{2.28}& 0.41\\ \hline
		\end{tabular}
	\end{table}

	\begin{table}
		\centering
		\caption{The heading angle RMSE values for Gaussian and uniform measurements.}
		\label{table:results1}
		\begin{tabular}{|c|c|c|c|c|}
			\hline
			\multirow{2}{*}{ }  & \multicolumn{2}{c|}{\textbf{\begin{tabular}[c]{@{}c@{}}Gaussian\\ Measurements\\ Heading\\ Angle RMSE {[}\textdegree{]}\end{tabular}}} & \multicolumn{2}{c|}{\textbf{\begin{tabular}[c]{@{}c@{}}Uniform \\Measurements\\ Heading \\ Angle RMSE {[}\textdegree{]}\end{tabular}}}  \\ \cline{2-5} 
			& \textbf{Mean} & \multicolumn{1}{c|}{\textbf{Std.}} & \textbf{Mean} & \multicolumn{1}{c|}{\textbf{Std.}} \\ \hline
			\textbf{Algorithm-1} &4.46&0.87 & 4.93 & 0.45\\ \hline
			\textbf{Algorithm-2} &59.98& 34.38& 60.15 & 34.00\\ \hline
			\textbf{VB}          &\textbf{3.93}& 0.41& \textbf{4.00} & 0.43\\ \hline
		\end{tabular}
	\end{table}

	\begin{figure*}
		\centering
		\includegraphics[width=\linewidth,trim={0 0 0 0},clip]{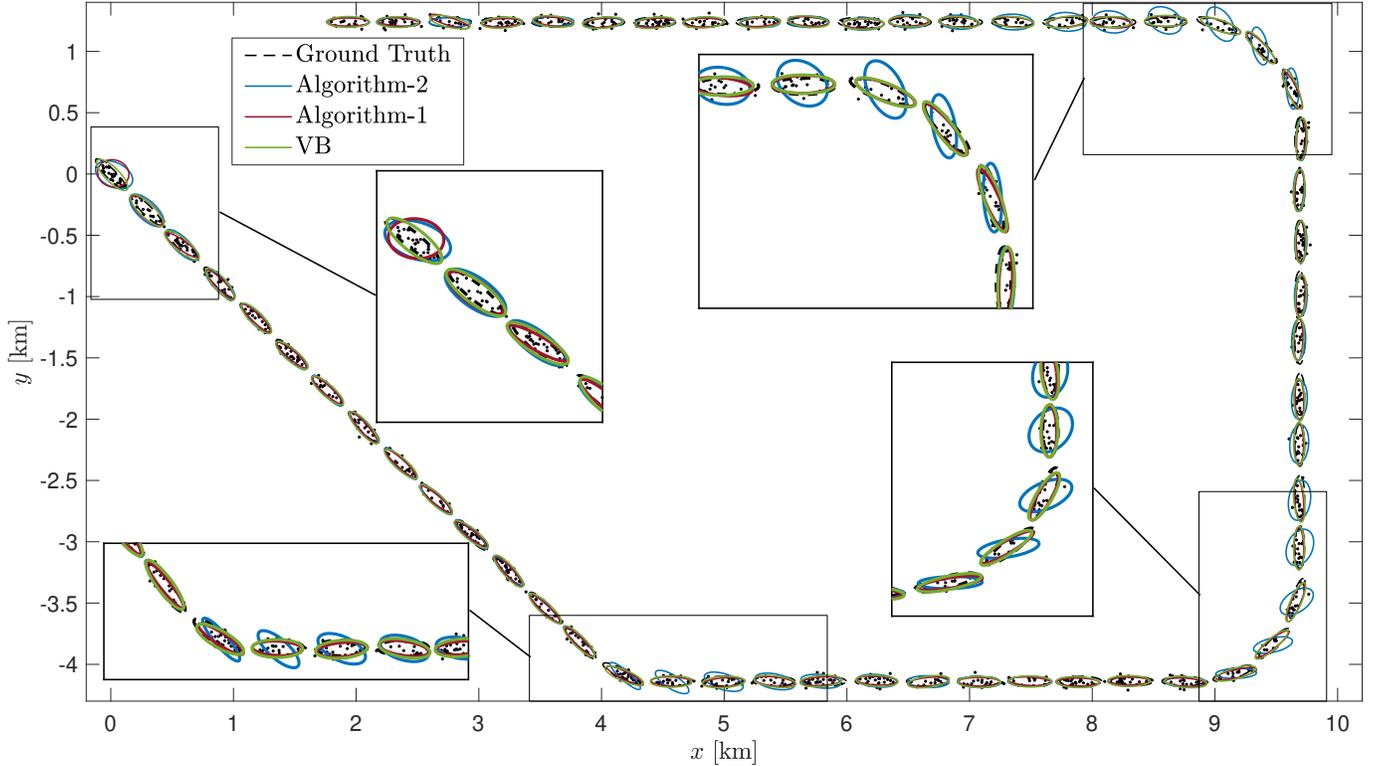}
		\captionof{figure}{An example MC run of the scenario in Section \ref{sec:yang_sce}}
		\label{fig:yangsimu}
	\end{figure*}
	
	\subsubsection{Experimental Trajectory}
	\label{sec:yang_sce}
	This experiment involves the scenario studied in \cite{Feldman,lan2012trackinga,yang,yang2017}.
	In this simulation, the trajectory composed of one $45$\textdegree{} and two $90$\textdegree{} turns pieced together with straight paths.
	The object of interest has unknown but fixed semi-axes lengths, and its orientation varies in time.
	The object starts its motion from the origin with a speed of 50 \si[per-mode=symbol]
	{\kilo\metre\per\hour}, which is fixed throughout the trajectory.
	The measurements are generated from a uniform distribution, and the number of the measurements is drawn from a Poisson distribution with an average of $20$ measurements per scan. 
	In addition to simulations performed in \cite{yang}, we will examine the performance of the algorithms with Gaussian distributed measurements.
	As in \cite{yang}, the prior mean and covariance matrix of the shape variables are selected to be $\hat{\textbf{p}}_0 = [\pi, \; 200, \; 90]^T$ and $C_0^p = \diag([1,\;70^2,\;70^2])$. 
	The process noise covariance matrix for the shape variables and kinematics are $C_p^w = \diag([0.1,\;1,\;1])$ and $C_r^w = \diag([100,\;100,\;1,\;1])$, respectively. 
	The measurement noise covariance matrix is, $R = \diag([400, \; 400])$. 
	In order to have a fair comparison, the prior mean values of the kinematic and shape variables for VB and Algorithm-2 are chosen to be the same as those of Algorithm-1.
	The shape variables for VB are selected to be $\mathbf{\alpha}_0^{1,2} = [5 \;5 ]$ and $\mathbf{\beta}_0^{1,2} =[400^2 \; 180^2]$.
	The degrees of freedom is $v_0 = 7$ for Algorithm-2.
	The scale matrix is initialized as $V_0 = \diag([400^2,\; 180^2])$.
	The initial mean of the kinematic state is set to $\hat{\textbf{x}}_0=[100  \; 100  \; 5  \; -8  \; \pi]^T$ for VB. The number of the variational iterations is 10.
	The initial mean of the kinematic state for the Algorithm-1 and Algorithm-2 is selected to be $\hat{\textbf{r}}_0 = \hat{\textbf{x}}_0(1:4)$. 
	We conducted 100 MC runs for each measurement distribution type.
	The GW distance and the orientation RMSE are presented in Table~\ref{table:yangsenaryouniform} and Table~\ref{table:yangsenaryogw}.
	An example MC run is depicted in Fig \ref{fig:yangsimu}.
	Algorithm-2 could not perform well during the turns because the method does not treat the orientation as a separate random variable and compensates the changes in the orientation by updating the extent estimate.  
	However, VB and Algorithm-1 are able to overcome this problem.
	The results show that the proposed approach, VB, provides better orientation, center, and extent estimates. 
	
	\begin{table}
		\centering
		\caption{The GW distance and heading angle RMSE values of the scenario in Section \ref{sec:yang_sce} when the measurements are uniformly distributed.}
		\label{table:yangsenaryouniform}
		\begin{tabular}{|c|c|c|c|c|c|c|}
			\hline
			\multirow{2}{*}{\textbf{}} & \textbf{\begin{tabular}[c]{@{}c@{}}GW Dist.\\ \nth{1} Term \\ {[}m\textsuperscript{2}{]}\end{tabular}} & \textbf{\begin{tabular}[c]{@{}c@{}}GW Dist.\\ \nth{2} Term \\{[}m\textsuperscript{2}{]}\end{tabular}} & \multicolumn{2}{c|}{\textbf{\begin{tabular}[c]{@{}c@{}}GW\\ Dist. \\{[}m{]}\end{tabular}}} & \multicolumn{2}{c|}{\textbf{\begin{tabular}[c]{@{}c@{}}Heading\\ Angle\\ RMSE \\{[}\textdegree{]}\end{tabular}}} \\ \cline{4-7}
			& & &\textbf{Mean} & \textbf{Std.} & \textbf{Mean} & \textbf{Std.} \\ \hline
			\textbf{Alg.-1} & 281.38 & 280.54 & 20.84 & 0.94 & 3.89 & 0.26 \\ \hline
			\textbf{Alg.-2} & 284.05 & 1436.05 & 32.94 & 082& 82.61 & 5.21\\ \hline
			\textbf{VB} & \textbf{270.74} & \textbf{203.01} & \textbf{19.83} & 0.89& \textbf{3.37}& 0.21\\ \hline
		\end{tabular}
	\end{table}
	
	\begin{table}
		\centering
		\caption{The GW distance and heading angle RMSE values of the scenario in Section \ref{sec:yang_sce} when the measurements are generated from a Gaussian distribution.}
		\label{table:yangsenaryogw}
		\begin{tabular}{|c|c|c|c|c|c|c|}
			\hline
			\multirow{2}{*}{\textbf{}} & \textbf{\begin{tabular}[c]{@{}c@{}}GW Dist.\\ \nth{1} Term \\ {[}m\textsuperscript{2}{]}\end{tabular}} & \textbf{\begin{tabular}[c]{@{}c@{}}GW Dist.\\ \nth{2} Term \\{[}m\textsuperscript{2}{]}\end{tabular}} & \multicolumn{2}{c|}{\textbf{\begin{tabular}[c]{@{}c@{}}GW\\ Dist. \\{[}m{]}\end{tabular}}} & \multicolumn{2}{c|}{\textbf{\begin{tabular}[c]{@{}c@{}}Heading\\ Angle\\ RMSE \\{[}\textdegree{]}\end{tabular}}} \\ \cline{4-7}
			& & &\textbf{Mean} & \textbf{Std.} & \textbf{Mean} & \textbf{Std.} \\ \hline
			\textbf{Alg.-1} & 884.75 & 244.15 & 29.17 &1.43 & 3.57 & 0.35\\ \hline
			\textbf{Alg.-2} & 826.32 & 1167.80 & 37.69 & 1.29 &82.66 & 5.08\\ \hline
			\textbf{VB} & \textbf{822.15} & \textbf{145.31} & \textbf{27.36} & 1.39& \textbf{2.94}& 0.20 \\ \hline
		\end{tabular}
	\end{table}
	
	\begin{figure}
		\centering
		\includegraphics[width=0.8\linewidth,trim={4cm 1cm 4cm 0cm},clip]{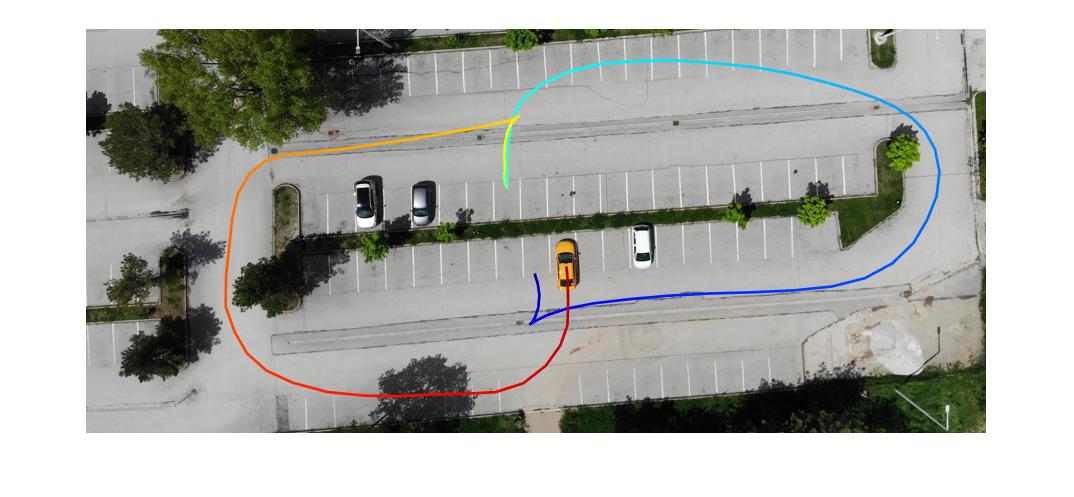}
		\captionof{figure}{The outline of the movement of the vehicle during the time-lapse. The vehicle starts from the dark blue colored parking spot; and follows the colored path until the red colored parking spot. In the figure, the last frame is shown.}
		\label{fig:kroki}
	\end{figure}

	\begin{figure*}
		\centering
		\includegraphics[width=\linewidth,trim={0 0 0 0},clip]{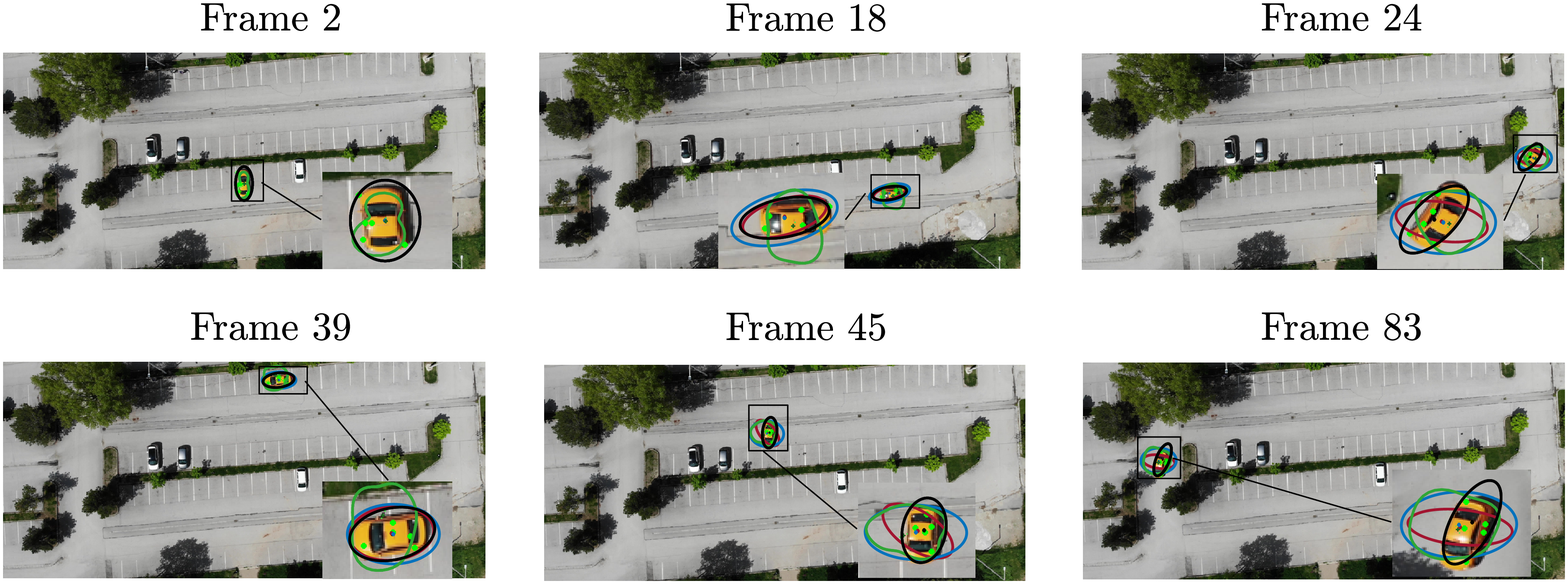}
		\captionof{figure}{A representative MC run of the real data experiment. The extent estimates of VB, Algorithm-1, Algorithm-2, and GP-ETT are shown in black, red, blue, and green lines, respectively. The measurements are represented with green dots.}
		\label{fig:kkmfig}
	\end{figure*}

	\subsection{Real Data Experiment} \label{subsec: realdata}
	In this section, the algorithms' capabilities are illustrated with real data.
	In addition to elliptical models, we compare the performance of another well-known ETT algorithm, namely the Gaussian process based extended target tracking (GP-ETT) algorithm \cite{EoGp1} to demonstrate the performance of the methods that do not rely on elliptical extent assumption in the scenario.
	
	The test data is collected in an urban area of Ankara.
	The test scenario involves a commercial vehicle moving in a parking lot while a steady aerial camera captures images of the surveillance region every second, i.e., $T=1s$. 
	In the scenario, a long sampling time is intentionally chosen to minimize the computational power consumption, thereby prolonging the air-time of the aerial camera in possible real-time applications. 
	The outline of the vehicle's trajectory is shown in Figure \ref{fig:kroki}.
	The colored-line indicates the trajectory followed by the mid-point of the vehicle.
	The scenario starts while the vehicle is parked in the parking area, indicated by the dark blue color.
	The vehicle leaves the parking area and follows the path shown in blue until it is parked in the parking area, which is indicated by the green color. 
	Then the vehicle performs a similar motion from the green-colored parking spot following the path to the red-colored parking spot. 
	
	Throughout the scenario, the captured images are processed for measurement extraction. Various feature extraction algorithms can be used to obtain measurements from the vehicle such as Harris corner detection \cite{harris}, Scale Invariant Feature Transform (SIFT) \cite{lowe2004distinctive}, Speeded Up Robust Features (SURF) \cite{bay2006surf} or similar. 
	In order to demonstrate that the algorithm can work with a wide range of feature extraction algorithms, we present a more general case, where the measurements are uniformly sampled from the vehicle's visible surface.
	The results obtained by using the features extracted by the Harris corner detector are also consistent with the results presented here, but they are not included in the manuscript because of the page limitations.
	
	As part of the image processing step, a segmentation is performed in every frame in the HSV color band to separate the yellow vehicle from the background. Following that, a median filter is used to reduce the number of clutters.
	Finally, the pixels that belong to the vehicle are sampled uniformly to obtain the measurements.
	The initial position of the vehicle is extracted from the first frame. 
	The initial velocity, on the other hand, is assumed to be unknown and assumed to be zero.
	Hence, the initial mean of the kinematic state vector is selected to be $\hat{\textbf{x}}_0 = [450\; 245\; 0\; 0\; \frac{\pi}{2}]^T$.
	The initial parameters of the algorithms are selected to match the prior means of the corresponding distributions.
	For this purpose, the initial shape parameters for VB is selected to be $\mathbf{\alpha}_0^{1,2} = [2 \; 2]^T$ and $\mathbf{\beta}_0^{1,2} = [250 \; 1000]^T$.
	The prior mean and covariance matrix of the shape variables  for Algorithm-1 is set to $\hat{\textbf{p}}_0 = [0 \;  250^{0.5} \; 1000^{0.5}]^T$ and $C^p_0 = \diag([1 , \; 100 , \;100 ])$, respectively.
	The degrees of freedom value and the initial scale matrix is set to $v_0 = 4$ and $V_0= \diag([250 , \; 1000])$ for Algorithm-2, respectively. 
	The process noise covariance matrix $Q$ is similar to the previous simulations, however $\sigma$ is taken as $4$, and $\sigma_\theta$ is $0.1$ for VB. The number of variational iterations is 10.
	The process noise covariance matrix for the shape variables for Algorithm-1 is set to $C^w_p = \diag([0.1, \; 10^{-3}, \; 10^{-3}])$.
	Finally, the measurement noise covariance matrix is taken as $R = \diag([1, \; 1])$.
	The parameters were optimized manually to obtain the best performances of the algorithms.
	
	The extent estimates corresponding to the frames $\{2,18,24,39,45,83\}$ are given in Figure \ref{fig:kkmfig}.
	These snapshots were chosen for the sake of a clearer illustration of the differences between the algorithms' performances, starting from the initial frames.
	At the beginning of the scenario, the vehicle stays immobile, and the algorithms are able to estimate the vehicle's extent satisfactorily (see: Frame $2$).
	
	Before the vehicle starts its movement, the extent estimate of the GP-ETT is more accurate and closer to the true extent of the target. On the other hand, the random matrix approaches are significantly advantageous throughout this scenario because they can fully exploit the prior information that the target's extent is close to an elliptical shape.
	GP-ETT algorithm aims at estimating the contours of objects having arbitrary shapes and its performance degrades when the measurements are originating from the surface of objects and the number of the surface measurements is low.
	The algorithm is essentially trying to solve a harder problem because it has more degrees of freedom to represent the unknown extent and a greater uncertainty to resolve compared to the ellipsoidal target tracking methods.
	
	When the vehicle is moving in a straight path, such as in Frame $18$ and Frame $39$, the performance of all random matrix based algorithms are satisfactory.
	However, when the vehicle performs a maneuver, as in Frame $24$, Frame $45$ and Frame $83$, VB shows superior performance in estimating the orientation of the vehicle.
	During the maneuvers, Algorithm-2 cannot estimate the extent accurately because it does not treat the heading angle as a separate random variable, and it tries to adapt to the changes in the orientation by updating the extent states.
	Algorithm-1 also struggles to find the correct orientation of the vehicle.
	However, VB can provide accurate estimates of the extent thanks to its iterative updates. 
	Note that VB and Algorithm-1 use the same process noise variance for the orientation.
	Since the vehicle is stable in the first couple of frames and the algorithms are able to estimate the extent accurately, increasing the variance values of the shape variables for Algorithm-1 does not improve the performance of estimating the extent further. 
	Additionally, if the variance values are increased too much, the extent estimates of Algorithm-1 tend to collapse to zero. We encountered a similar problem while tuning Algorithm-1 in the simulation scenarios. We report one example of such behavior in a single measurement update in Appendix \ref{appen:collapse} for interested readers.
	
	\section{Conclusion and Discussion}
	\label{sec:Conclusion}
	ETT involves tracking objects that generate multiple measurements per scan. 
	In most ETT applications, the orientation of the extended targets changes in time. 
	In standard RM based ETT methods, this phenomenon is addressed by a forgetting factor, which aims at forgetting the accumulated information.
	In this work, we proposed a novel approach for extended target tracking that is capable of simultaneously estimating the kinematic, extent, and orientation states of an extended target. 
	We use the variational Bayes technique for inference and define appropriate priors for the unknown state variables that can accurately model the changes in the extended targets' orientation. The performance and capabilities of the algorithm are demonstrated through simulations and real data experiments.
	Experimental results on simulations and real data demonstrate that the proposed method significantly improves the tracking performance, as well as the accuracy in estimating the orientation and the shape of the object compared to the state-of-the-art methods.
	
	It is also worth mentioning that variational Bayes approaches resort to factorized distributions, which lose the correlation structure in the posterior density. An algorithm that does not neglect correlation terms may provide better estimation performance than such variational methods. In our experience (as illustrated in benchmark scenarios and real data experiments in Sections~\ref{sec:single_meas}-\ref{sec:results}), the iterative optimization structure provided by the variational inference framework outperforms the alternative solutions by overcoming the disadvantages associated with the factorized approximation. However, improved performance can be achieved by further exploiting the correlation structure in the true posterior.
	
	The model and the technique we use might also be applicable to estimation problems other than extended target tracking applications, which involve dynamic elliptical representations or unknown covariance matrices with similarly structured uncertainty.
	These problems may include, but are not limited to, obtaining elliptical bounds in power systems \cite{qing2013extended, chen2019distributed}, estimating ellipsoid sets containing target states over sensor networks \cite{ding2019set} or spectrum representation in speech processing \cite{DynamicSpeech, carevic2016two}.
	
	\ifCLASSOPTIONcaptionsoff
	\newpage
	\fi
	
	\bibliographystyle{IEEEtran}
	\bibliography{mybib}
	
	\begin{appendices}
		\section{Calculation of $\delta$ and $\Delta$}
		\label{appen:mM}
		In Section \ref{eq:tet_update}, we introduced the update formulas for the orientation distribution as below.
		\begin{align}
		\hat{\theta}^{(\ell+1)}_{k|k} &= \Theta_{k|k}^{(\ell+1)}\big( \Theta_{k|k-1}^{-1}\hat{\theta}_{k|k-1}+\delta \big),\\
		\Theta_{k|k}^{(\ell+1)} &= \big (\Theta_{k|k-1}^{-1} + \Delta \big)^{-1},
		\end{align}
		Here, the variables $\delta$ and $\Delta$ are derived from the expectation in \eqref{eq:exp_tet}.
		\begin{align}
		\label{eq:exp_tet}
		&\frac{-1}{2} \sum_{j=1}^{m_k} \E_{\backslash\theta_k}\left[(a-b\theta_k)^T(sX)^{-1}(a-b\theta_k)\right], 
		\end{align}
		where $\delta$ and $\Delta$ are denoted as
		\begin{align}
		\label{eq:m}
		&\delta \triangleq \sum_{j=1}^{m_k}\tr\left[\E_{q_{X}^{(\ell)}}\left[(sX_k)^{-1}\right] \E_{q_{\mathbf{x}}^{(\ell)},q_{\textbf{Z}}^{(\ell)}}\left[ab^T\right]\right], \\
		\label{eq:M}
		&\Delta \triangleq \sum_{j=1}^{m_k}\tr\left[\E_{q_{X}^{(\ell)}}\left[(sX_k)^{-1}\right] \E_{q_{\mathbf{x}}^{(\ell)},q_{\textbf{Z}}^{(\ell)}}\left[bb^T\right]\right], \\
		\text{and}&\nonumber \\
		&a = (T_{\hat{\theta}^{(\ell)}_{k|k}})^T (\mathbf{z}_k^j-H\mathbf{x}_k)-(T_{\hat{\theta}^{(\ell)}_{k|k}}^{\prime})^T(\mathbf{z}_k^j-H\mathbf{x}_k)\hat{\theta}_{k|k}^{(\ell)},\\
		&b = -(T_{\hat{\theta}^{(\ell)}_{k|k}}^{\prime})^T(\mathbf{z}_k^j-H\mathbf{x}_k).
		\end{align} 
		When the $a$ and $b$ variables are substituted into \eqref{eq:m} and \eqref{eq:M}, we obtain the $\delta$ and $\Delta$ variables as
		\begin{align}
		\delta &=\sum_{j=1}^{m_k} \tr\bigg[\overline{sX_k^{-1}}(T_{\hat{\theta}^{(\ell)}_{k|k}}^{\prime})^T\overline{\big(\mathbf{z}_k^j-H\mathbf{x}_k\big)\big(\cdot\big)^T}(T_{\hat{\theta}^{(\ell)}_{k|k}}^{\prime})\hat{\theta}^{(\ell)}_{k|k}   \bigg] \nextline -\tr\bigg[\overline{sX_k^{-1}}T_{\hat{\theta}^{(\ell)}_{k|k}}^T\overline{\big(\mathbf{z}_k^j-H\mathbf{x}_k\big)\big(\cdot\big)^T}(T_{\hat{\theta}^{(\ell)}_{k|k}}^{\prime})   \bigg],\\
		\Delta &= \sum_{j=1}^{m_k} \tr\bigg[\overline{sX_k^{-1}}(T_{\hat{\theta}^{(\ell)}_{k|k}}^{\prime})^T\overline{\big(\mathbf{z}_k^j-H\mathbf{x}_k\big)\big(\cdot\big)^T}(T_{\hat{\theta}^{(\ell)}_{k|k}}^{\prime})\bigg],
		\end{align}
		
		\noindent where $\overline{\big(\mathbf{z}_k^j-H\mathbf{x}_k\big)\big(\cdot\big)^T} = \E_{q_{\mathbf{x}}^{(\ell)},q_{\textbf{Z}}^{(\ell)}}\left[\big(\mathbf{z}_k^j-H\mathbf{x}_k\big)\big(\cdot\big)^T\right]$, and $\overline{sX_k^{-1}} = \E_{q_{X}^{(\ell)}}\left[(sX_k)^{-1}\right]$.
		
		\section{Proof of Lemma~1}
		\label{appen:proof}
		In this section we will give the proof of \textit{Lemma}~1 to calculate $\E_{q_{\theta}^{(\ell)}} \left[( s T_{\theta_k} M T_{\theta_k}^T)^{-1} \right]$.
		In the formulation, we first multiply the matrices inside the expectation. 
		Then, the expectation of each entry of the resultant matrix is taken.
		First, notice that,
		\begin{align}
		\label{eq:matrix}
		\big(  T_{\theta_k} M T_{\theta_k}^T\big)^{-1}  = T_{\theta_k} M^{-1} T_{\theta_k}^T.
		\end{align}
		Given 
		\begin{align*}
		M^{-1} = \begin{bmatrix}
		m_{11} & m_{12}\\ 
		m_{21} & m_{22}
		\end{bmatrix},
		\end{align*}
		the expression whose expectation has to be taken becomes
		\begin{subequations}\label{eq:expressions}
			\begin{align}
			\Lambda &= T_{\theta_k} M^{-1} T_{\theta_k}^T \nonumber \\ &= \begin{bmatrix} \cos(\theta_k) & -\sin(\theta_k)\\ \sin(\theta_k) & \cos(\theta_k) \end{bmatrix} \begin{bmatrix}
			m_{11} & m_{12}\\ 
			m_{21} & m_{22}
			\end{bmatrix} \nextline \times \begin{bmatrix} \cos(\theta_k) & \sin(\theta_k)\\ -\sin(\theta_k) & \cos(\theta_k) \end{bmatrix},\\ 
			\Lambda(1,1) &=  m_{11}\cos^2(\theta_k)+m_{22}\sin^2(\theta_k)\nextline-(m_{12}+m_{21})\cos(\theta_k)\sin(\theta_k),\\
			\Lambda(1,2) &= m_{12}\cos^2(\theta_k)-m_{21}\sin^2(\theta_k)\nextline+( m_{11}-m_{22})\cos(\theta_k)\sin(\theta_k), \\
			\Lambda(2,1) &= m_{21}\cos^2(\theta_k)-m_{12}\sin^2(\theta_k)\nextline+( m_{11}-m_{22})\cos(\theta_k)\sin(\theta_k), \\
			\Lambda(2,2) &= m_{22}\cos^2(\theta_k)+ m_{11}\sin^2(\theta_k)\nextline+(m_{12}+m_{21})\cos(\theta_k)\sin(\theta_k). 
			\end{align}
		\end{subequations}
		Now the following trigonometric transformations are utilized and the expectations are taken with respect to the resulting expressions.
		\begin{subequations} \label{eq:equality}
			\begin{align}
			\cos^2(\theta_k) &= \frac{1+\cos(2\theta_k)}{2}, \quad 
			\sin^2(\theta_k) = \frac{1-\cos(2\theta_k)}{2}, \\
			\cos(\theta_k)\sin(\theta_k) &= \frac{\sin(2\theta_k)}{2}, \\
			\E_{q_{\theta}^{(\ell)}}\left[\cos(2\theta_k)\right] &= \cos(2\hat{\theta}_{k|k}^{(\ell)})\exp(-2\Theta_{k|k}^{(\ell)})), \\
			\E_{q_{\theta}^{(\ell)}}\left[\sin(2\theta_k)\right] &= \sin(2\hat{\theta}_{k|k}^{(\ell)})\exp(-2\Theta_{k|k}^{(\ell)})).
			\end{align}
		\end{subequations}
		By substituting the expressions in \eqref{eq:expressions} with the corresponding equalities given in \eqref{eq:equality} \textit{Lemma}~1 is obtained.
		
		\section{The parameters of the experiment in section \ref{sec:single_meas}}
		\label{sec:params}
		In this section, the parameters of the experiment given in Section \ref{sec:single_meas} are summarized.
		The prior shape estimate is apart from the true center location of the target by $(20,\; 20)$ units in the 2D coordinate frame.
		The extent of the ground truth object is parametrized as diag$([6, \; 0.5])$ with $0\degree$ orientation.
		The mean of the prior extent ellipse parameters for both methods are equal to the parameters of the ground truth extent.
		The prior shape parameters of the proposed algorithm are selected to be $\mathbf{\alpha}_0^{1,2} = [101 \; 101]^T$ and $\mathbf{\beta}_0^{1,2} = [600 \; 50]^T$. 
		The prior mean of the orientation variable is taken as $\hat{\theta}_0 = \frac{\pi}{2}$.
		To have a reasonable comparison, the mean vector and the variance of the shape parameters for the EKF approach are selected to match with those of the VB algorithm.
		The prior kinematic state covariance matrix is $P_0 = \diag([300,\;300,\;1,\;1,\;\frac{\pi}{2}])$.
		The measurement noise covariance matrix is $R = \diag([1, \; 1])$.
		In this experiment, the number of measurements is $10$, and the measurements are generated according to a Gaussian distribution.
		However, the trials with the uniformly distributed measurements yield similar results.

		\section{An example for the collapsing extent estimates}
		\label{appen:collapse}
\begin{figure}[H]
			\centering
			\includegraphics[scale=0.6,trim={0 0 0 0},clip]{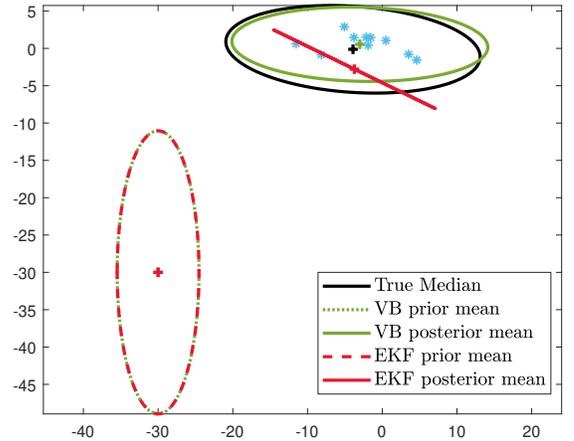}
			\captionof{figure}{The visualization of the collapsing behavior of Algorithm-1. Blue stars represent the measurements, the solid green and red lines stand for the posterior means of the VB and EKF updates, respectively. The solid black line indicates the median of the true posterior, which is computed by using 1 million Monte Carlo samples.}
			\label{fig:collapse}
		\end{figure}

		Here we repeated the simulation in Section \ref{sec:single_meas} with the following set of parameters:
		
		\noindent 
		For VB, the prior mean of the target's kinematic state is $\hat{\textbf{x}}_0=[-30 \; -30 \; 1 \; 1 \; \frac{\pi}{2}]^T$.
		The prior kinematic state covariance matrix is $P_0 = \diag([300,\;300,\;1,\;1,\;\frac{\pi}{2}])$.
		The shape parameters for VB is $\mathbf{\alpha}^{1,2} = [11 \; 11]^T$ and $\mathbf{\beta}^{1,2} = [600 \; 50]^T$.
		For Algorithm-1, the prior mean vector and covariance matrix of the kinematic state is $\hat{\textbf{r}}_0 = \hat{\textbf{x}}_0(1:4)$ and $C_0^r = \diag([300,\;300,\;1,\;1])$, respectively.
		The prior mean and variance of the extent parameters are the same for both algorithms.
		The number of measurements generated from the target is $10$.
		The measurement noise covariance matrix is $R = \diag([1, \; 1])$.
		A single measurement update of VB and Algorithm-1 is visualized in Figure \ref{fig:collapse}.
		
	\end{appendices}
	
	\begin{IEEEbiography}[{%
		\includegraphics[width=1in,height=1.25in,clip,keepaspectratio]{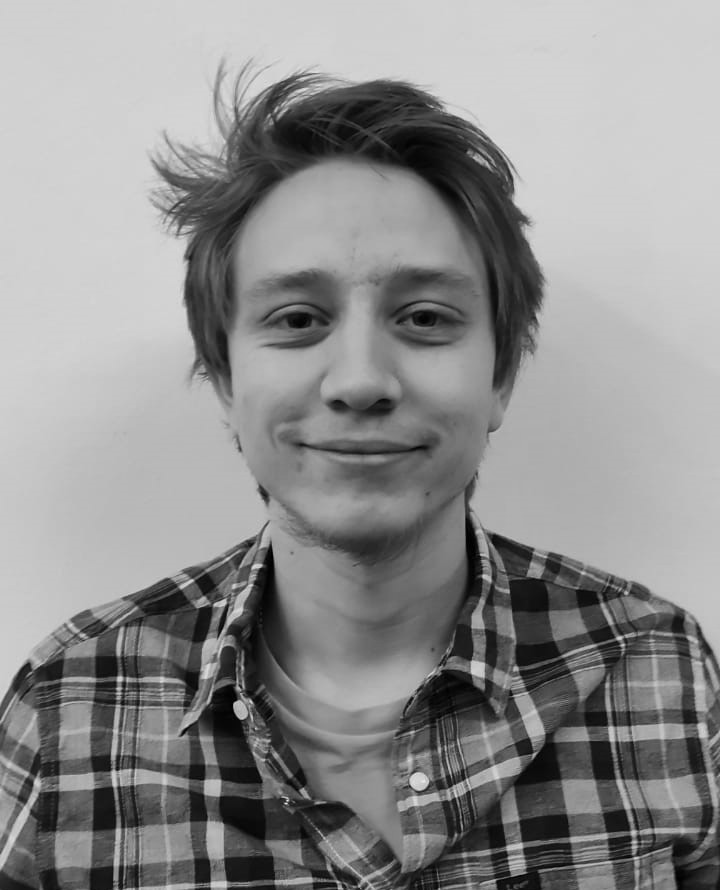}
	}]{Barkın Tuncer}(S'20)
	Barkın Tuncer received his B.Sc. degree in Electrical and Electronics Engineering from Middle East Technical University (METU), Ankara, Turkey, 2019. Since 2019, he has been working toward the M.Sc. degree in the Robotics specialization area at METU while working as a teaching assistant. His research interest includes extended target tracking, statistical signal processing, Bayesian inference, machine learning, and deep learning. 
\end{IEEEbiography}

	\begin{IEEEbiography}[{%
	\includegraphics[width=1in,height=1.25in,clip,keepaspectratio]{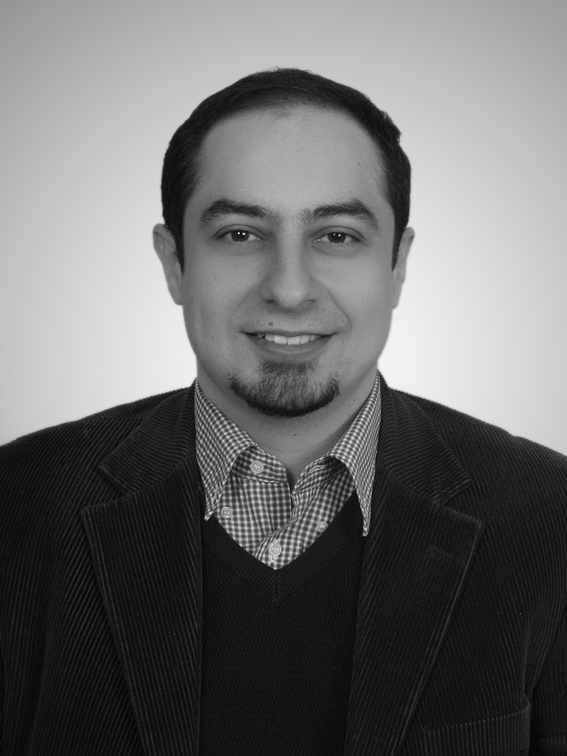}
}]{Emre {\"O}zkan}(S'04 - M'10)
received his B.Sc. and PhD degrees in Electrical and Electronics Engineering from Middle East Technical University (METU), Ankara, Turkey, in 2009. Between 2009 and 2011, he worked as a postdoctoral associate in the Division of Automatic Control, Department of Electrical Engineering, Link\"oping University, Sweden. Between 2011 and 2015 he worked as an Assistant Professor in the same group. In 2015, he worked as a visiting researcher at Signal Processing and Communications Laboratory at the University of Cambridge. Currently, he is an Assistant Professor at METU. His research focuses on target tracking, statistical signal processing, estimation theory, sensor fusion, Monte Carlo methods and Bayesian inference.
\end{IEEEbiography}

\end{document}